\documentclass{ieeeaccess}
\usepackage{cite}
\usepackage{amsmath,amssymb,amsfonts}
\usepackage{algorithmic}
\usepackage{graphicx}
\usepackage{textcomp}
\usepackage{mathtools, bm, bbm}
\usepackage{algorithmic}
\usepackage{comment} 
\usepackage{mathrsfs}
\usepackage[ruled,vlined]{algorithm2e}
\usepackage{enumitem}
\usepackage{color}
\usepackage{booktabs}
\usepackage{multirow} 
\usepackage[switch]{lineno}
\usepackage[linkcolor=red, citecolor=black,colorlinks=true]{hyperref}
\usepackage{array}
\usepackage{url}
\def\BibTeX{{\rm B\kern-.05em{\sc i\kern-.025em b}\kern-.08em
    T\kern-.1667em\lower.7ex\hbox{E}\kern-.125emX}}

\newcommand{\etal}{\textit{et al.}}
\newcommand{\ie}{\textit{i.e., }}
\newcommand{\eg}{\textit{e.g., }}
\definecolor{black}{rgb}{0.0, 0.0, 0.0}
\newcommand{\hl}[1]{\textcolor{black}{#1}} 
\newcommand{\hlt}[1]{\textcolor{black}{#1}} 
\newcommand{\hltt}[1]{\textcolor{black}{#1}} 
\definecolor{orange}{rgb}{1.0, 0.5, 0.0}

\newtheorem{theorem}{Lemma}
\begin{document}
\title{Fast Gravitational Approach for Rigid Point Set Registration with Ordinary Differential Equations} 
\author{
\uppercase{SK AZIZ ALI\authorrefmark{1,}\authorrefmark{2}}, $\,$
\uppercase{KEREM KAHRAMAN\authorrefmark{2}}, $\,$
\uppercase{Christian Theobalt\authorrefmark{3}}, $\,$
\uppercase{Didier Stricker\authorrefmark{1,}\authorrefmark{2}}, $\,$
\uppercase{AND Vladislav Golyanik\authorrefmark{3}}
}
\address[1]{Dept. of Computer Science, Technische Universit\"at Kaiserslautern, Germany}
\address[2]{German Research Center for Artificial Intelligence, Kaiserslautern (DFKI GmbH), Augmented Vision Group}
\address[3]{Max Planck Institute for Informatics, Saarland Informatics Campus}
\tfootnote{This work was partially supported by the project VIDETE (01IW18002) of the German Federal Ministry of Education and Research (BMBF) and by the ERC Consolidator Grant 4DReply (770784). 
}

\corresp{Corresponding author: Sk Aziz Ali (e-mail: sk\_aziz.Ali@dfki.de) and Vladislav Golyanik (e-mail: golyanik@mpi-inf.mpg.de)}

\begin{abstract}
This article introduces a new physics-based method for rigid point set alignment called Fast Gravitational Approach (FGA). In FGA, the source and target point sets are interpreted as rigid particle swarms with masses interacting in a globally multiply-linked manner while moving in a simulated gravitational force field. The optimal alignment is obtained by explicit modeling of forces acting on the particles as well as their velocities and displacements with second-order ordinary differential equations of n-body motion. Additional alignment cues can be integrated into FGA through particle masses. We propose a smooth-particle mass function for point mass initialization, which improves robustness to noise and structural discontinuities. To avoid the quadratic complexity of all-to-all point interactions, we adapt a Barnes-Hut tree for accelerated force computation and achieve quasilinear complexity. We show that the new method class has characteristics not found in previous alignment methods such as efficient handling of partial overlaps, inhomogeneous sampling densities, and coping with large point clouds with reduced runtime compared to the state of the art. Experiments show that our method performs on par with or outperforms all compared competing deep-learning-based and general-purpose techniques (which do not take training data) in resolving transformations for LiDAR data and gains state-of-the-art accuracy and speed when coping with different data.
\end{abstract}

\maketitle
\section{Introduction}\label{sec:introduction}
\PARstart{R}{igid} point set registration (RPSR) is essential in many computer vision and computer graphics tasks such as camera pose estimation~\cite{marchand2015pose}, 3D reconstruction~\cite{SALVI2007578}, CAD modeling, object tracking and simultaneous localization and mapping~\cite{KinextFusion_Newcombe,Whelan-2015-5935} and autonomous vehicle control~\cite{Zhang-2016-110808}, to name a few. 
Suppose we would like to merge partial 3D scans obtained by structured light into a single and complete 3D reconstruction of a scene, identify a pre-defined pattern in the 3D data or estimate the trajectory of the sensor delivering 3D point cloud measurements. 
All of these tasks can be addressed by a robust RPSR approach which can cope with partially 
overlapping and noisy data. 

The objective of \textit{pairwise} RPSR is, given a pair of unordered sets of points generally in 2D, 3D, or higher-dimensional space, to find optimal rigid transformation parameters (\eg $6$DOF in 3D, rotation $\mathbf{R} \in SO(3)$ and translation $\mathbf{t} \in \mathbb{R}^3$) aligning the \textit{template} point set to the fixed \textit{reference} point set. 
One of the earliest method classes --- iterative closest points (ICP) \cite{BeslMcKay1992,ChenMedioni1992} --- is still among the most widely-used techniques nowadays, due to its simplicity and speed. In ICP, the problem of RPSR is converted to transformation estimation between points with known point correspondences. In each iteration, correspondences are selected according to the nearest neighbor rule. Nevertheless, ICP is not the ideal choice in many challenging scenarios with noise, partial overlaps or missing entries in the data, due to its inherent sensitivity to all these disturbing effects and the deterministic correspondence selection rule. To tackle these difficulties, many other techniques for RPSR were subsequently proposed over the last decades~\cite{5713477-Survey, Pomerleau2013}. 

\begin{figure*}[!ht] 
\begin{center}
\includegraphics[width=0.99\linewidth]{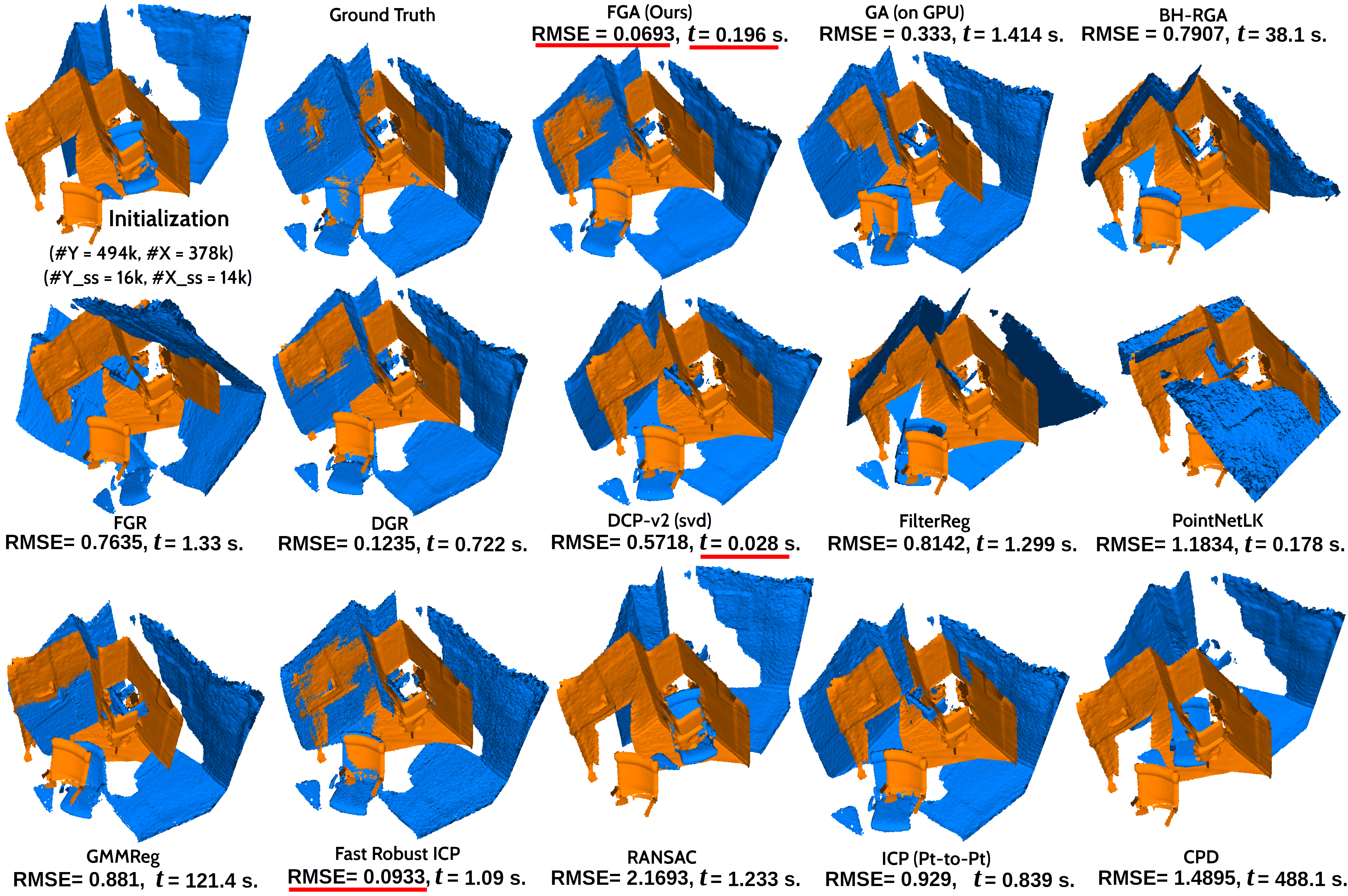} 
\end{center}
\vspace{-0.25cm}
\caption{A comparison between several standout rigid registration methods (state-of-the-art) on a randomly selected 
pair of low-overlapping point clouds from popular 3DMatch~\cite{zeng20163dmatch} dataset. $\#\mathbf{Y}$ and $\#\mathbf{X}$ are the initial point sizes of the source and target scans.
Due to the computational constraint, the corresponding subsampled versions of the inputs (with point sizes $\#\mathbf{Y}_{ss}$ and $\#\mathbf{X}_{ss}$) are used in the tests.
The comparison shows alignment accuracy as root mean squared error (RMSE) of distances between actual and transformed source, and runtime of each tested method. 
The RMSE metric, which is mesured as 
$(\frac{1}{\#Y}\sum_{i=1}^{\#Y}\lVert\mathbf{R}_{gt}\mathbf{Y} + \mathbf{t}_{gt} - \mathbf{R}^{*}\mathbf{Y} - \mathbf{t}^{*}  \rVert_{2}^{2})^{1/2}$, defines the alignment accuracy based on the ground-truth transformation ($\mathbf{R}_{gt}, \mathbf{t}_{gt}$) and estimated transformation ($\mathbf{R}^{*}, \mathbf{t}^{*}$). The red underline markers highlight the best case scenarios. Our method FGA is compuationally fast and robust compared to  GA~\cite{Golyanik2016GravitationalAF}, BHRGA~\cite{BHRGA2019}, FGR~\cite{FGRECCV16}, DGR~\cite{choy2020deep}, DCP-v2~\cite{wang2019deep}, FilterReg~\cite{Gao2019}, PointNetLK~\cite{AokiGSL19}, Fast Robust ICP~\cite{zhang2021FRICP}, point-to-point ICP~\cite{BeslMcKay1992}, RANSAC~\cite{Rusu2009}, GMMReg~\cite{GMMReg1544863} and CPD~\cite{MyronenkoSong2010}.} 
\label{fig_sim}
\end{figure*}

In recent times, RPSR methods relying on physical analogies~\cite{Golyanik2016GravitationalAF, Deng2014, Jauer2019, BHRGA2019} are emerging. They offer an alternative perspective to the problem and can often successfully handle cases which are difficult for other algorithmic classes. Physics-based methods have been successful in many domains of computer vision~\cite{LaiYung1998,Zhang2019,Nixon2009} and have multiple advantages over other method classes. Moreover, by using physical principles, we have access to a large volume of research on computational physics. For instance, we can borrow data structures and acceleration techniques which were successfully applied in numerical simulations~\cite{GREENGARD1987325,spurzem2009accelerating}.

Although RPSR is a well-studied research area, we believe that further advances are possible here with physics-based techniques. With new sensors (\eg LiDAR), the spatial properties like sampling accuracy and density or physical properties like light-reflectance and color-consistency of the point clouds are considerable factors for the current alignment techniques. Thus, automotive applications require real-time methods for aligning large, partially overlapping data with outliers and inhomogeneous point densities. 

\subsection{Contributions} 
In this article, we propose a new physics-inspired approach to rigid point set alignment --- Fast Gravitational Approach (FGA, Sec.~\ref{sec:ProposedGA_For_RPSR}) --- which is our core contribution.
In FGA, point sets are interpreted are particle swarms with masses moving under the simulated gravitational force field induced by the reference as depicted in Fig.~\ref{fig_sim}. The consecutive states of the template are obtained by explicit modeling of Newtonian particles dynamics and solving for displacements and velocities of the particles with second-order ordinary differential equations (ODEs). In contrast to methods based on correspondences selection and filtering \cite{Rusu2009, FGRECCV16, 2020arXiv200107715Y}, FGA is a correspondence-free approach, \textit{i.e.,} its energy function is defined in terms of interactions between all template and reference points. Thus, our method is globally-multiply linked and has properties not found in other related algorithmic classes (\textit{e.g.,} high robustness to noise, versatile applicability of 
point masses and the fact that the locally-optimal alignment is reached when the gravitational potential energy of the system is locally-minimal).

This article is partially based on our published conference paper~\cite{Golyanik2016GravitationalAF}, while featuring further contributions towards the accuracy gain, computational complexity reduction, handling inhomogeneity in point clouds and feature-based boundary conditions for tackling partially overlapping data. To summarize: 
\begin{itemize}
 \item Besides the general settings of particle interactions and dynamics in \cite{Golyanik2016GravitationalAF} (Secs.~\ref{subsec:NotationsAndDefinition}--\ref{subsubsec:Rigidity_Constraing_GA}), we propose an acceleration technique with point clustering, \ie Barnes-Hut (BH) tree~\cite{1986Natur324446B} (Secs.~\ref{subsec:AccPolicy}--\ref{subsec:BH-Tree Force}) which reduces the algorithm's quadratic computational complexity to quasilinear. Especially for large point sets as those arising in automotive and augmented reality applications, FGA without acceleration can become prohibitive. Subsampling can alleviate the problem but leads to data loss (especially of high-frequency details) as, ideally, one would like to use all available data. In contrast, our acceleration technique preserves the explicit influence of all available data points as well as the globally multiply-linked point interactions.
 \vspace{0.15cm}
 \item Next, we show that particle masses in FGA can be initialized using different types of boundary conditions such as prior correspondences and feature-based alignment cues 
 (Sec.~\ref{subsec:BoundaryConditions}). Thus, we propose a \textit{normalized intrinsic volume} (NIV) measure per point to be assigned as their mass. This is an effective weighting scheme which smoothly balances the inhomogeneous point sampling density. Similarly, if some matches are given, radially symmetric weights can be assigned to the masses via a radial basis function (RBF)~\cite{doi:10.1029/JB076i008p01905}. In the case of partially overlapping data with some known prior correspondences, the Hadamard product of RBF and NIV values as the assigned masses makes the method robust and addresses the local minima issue. Other point features like Fast Point Feature Histogram (FPFH)~\cite{Rusu2009} are also adaptable for the point mass initialization. 
 \vspace{0.15cm}
 \item As it has been recently shown, resolving the uniform scale difference, \textit{i.e.,} 
 the seventh DOF is not possible while using the globally multiply-linked interaction policy~\cite{proof_singularity_2019}. Thus, in 3D 6DOF pose estimation, we perform an extensive evaluation of FGA against multiple widely-used and state-of-the-art rigid point set alignment methods and show applications of FGA to LiDAR-based odometry~\cite{li2019net} and completions of real RGB-D scans~\cite{sturm12iros} (Sec.~\ref{sec:Experiments}). FGA outperforms general-purpose RPSR methods in scenarios with large amounts of noise and partial overlaps. We can cope with large samples originating from the modern LiDAR sensors --- containing ${\sim}400k$ points each --- and can support point cloud based odometry at 
 ${\sim}1.5$ pairs per second with our GPU version of FGA. In contrast, other competing methods, to the best of our knowledge, cannot cope with such large data in such short time. 
\end{itemize}

\subsection{Structure of the Article} 
The rest of the article is organized as follows. In Sec.~\ref{sec:N_BodySimulation}, we first review the classical $\mathtt{n}$-body problem and show how it can be adapted for RPSR. 
FGA is introduced in Sec.~\ref{sec:ProposedGA_For_RPSR}. We define our gravitational model with second-order ODEs and simulation rules in Secs.~\ref{subsec:NotationsAndDefinition}--\ref{subsec:AccPolicy}. Details on BH tree building and integration into FGA for accelerated alignment can be found in Sec.~\ref{subsec:BH-Tree Force}. Embedding boundary conditions through masses is elaborated in Sec.~\ref{subsec:BoundaryConditions}. 
FGA is then evaluated against multiple RPSR methods in Sec.~\ref{sec:Experiments}, followed by discussion and conclusive remarks.

\vspace{-0.15cm}
\section{Related Work}\label{sec:related_work} 
Related work on point set alignment is vast and can be reviewed from different perspectives. 
We classify the methods according to whether they alternate between the correspondence 
search and transformation estimation (Sec.~\ref{ssec:ICP}), rely on probabilistic modeling 
(Sec.~\ref{ssec:probabilistic_methods}) of input data, are deep-learning-based 
(Sec.~\ref{ssec:deepLearning_based_methods}) or are physics-based 
(Sec.~\ref{ssec:physics_based_methods}).  
 
\vspace{-0.15cm}
\subsection{From Transformation Estimation to ICP}\label{ssec:ICP} 
Some approaches \cite{Rusu2009, FGRECCV16, Yew_2018_ECCV, 2020arXiv200107715Y} first extract a sparse set of descriptive key points from point sets~\cite{Rusu2008, Zhong2009} and then find optimal alignment parameters with a transformation estimation approach~\cite{Kabsch1976, Horn87, Arun1987, Horn88, Umeyama1991}. This policy does not use all available points and often leads to coarse alignments but, on the other hand, can result in a significantly improved initialization for other RPSR approaches \cite{Rusu2008}. Iterative Closest Point (ICP), pioneered by Besl and McKay \cite{BeslMcKay1992} as well as Chen and Medioni \cite{ChenMedioni1992}, is an approach alternating between correspondence search and transformation estimation. 

Various modifications of ICP have been introduced over the years \cite{Greenspan2003, Elseberg12, Fitzgibbon01c, Rusinkiewicz2001} to tackle its local minima trapping issue. 
Segal and coworkers \cite{Segal2009} extended the classical ICP with probabilistic transformation estimation. In \cite{Pomerleau2013}, a comprehensive overview of ICP variants is available.

\vspace{-0.2cm}
\subsection{Probabilistic Methods}\label{ssec:probabilistic_methods} 
Probabilistic approaches assign a probability of being a valid correspondence to point 
pairs~\cite{ChuiRangarajan2000, GMMReg1544863, MyronenkoSong2010}. 
Chui and Rangarajan \cite{ChuiRangarajan2000} interpret point set alignment as a mixture density estimation problem. Their Mixture Point Matching (MPM) approach iteratively updates probabilistic correspondences and the transformation with annealing and Expectation-Maximization (EM) schemes, respectively. Coherent Point Drift (CPD) \cite{MyronenkoSong2010} models the template as a Gaussian Mixture Model (GMM) which is fit to the reference interpreted as data points. 
FilterReg \cite{Gao2019} pursues an alternative approach, \textit{i.e.,} the reference 
induces a GMM. This results in a simpler, faster and --- in some scenarios --- a more accurate 
algorithm compared to CPD. In contrast to \cite{MyronenkoSong2010, Gao2019}, GMM Registration (GMMReg) \cite{GMMReg1544863} interprets both point sets as GMM, and point set alignment 
is posed as mixture density alignment. The approach of Tsin and Kanade \cite{TsinKanade2004} finds a configuration with the highest correlation between point sets leading to the optimal alignment. 
Eckart~\textit{et al.}~align inhomogeneous point clouds with hierarchical GMM \cite{Eckart2018HGMRHG}. Several approaches additionally use alignment cues as prior correspondences or colors of point clouds \cite{Golyanik2016rECPD, Danelljan2016, Park2017, SavalCalvo2018}. Our formulation integrates prior correspondences and point features by mapping them to point masses. 

\vspace{-0.2cm}
\subsection{Deep Learning Approaches}\label{ssec:deepLearning_based_methods} 
Multiple techniques with deep neural networks (DNN) for point cloud processing tasks 
(\textit{e.g.,} classification and segmentation~\cite{qi2017pointnet, qi2017pointnet++} 
or shape matching~\cite{gojcic2019perfect, ShimadaDispVoxNets2019, Donati_2020_CVPR}) 
have been recently proposed. RPSR approaches using DNNs have also appeared recently in 
numbers~\cite{Elbaz20173DPC, AokiGSL19, wang2019deep, wang2019prnet, lu2019deepvcp, gross2019alignnet, yuan2020deepgmr}. Most of them~\cite{AokiGSL19, wang2019deep, wang2019prnet, lu2019deepvcp} utilize PointNet~\cite{qi2017pointnet} as a deep feature extractor and feature matching layers for estimating rigid transformations. In contrast, Deep Global Registration (DGR)~\cite{choy2020deep} --- which is a data-driven version of Fast Global Registration (FGR)~\cite{FGRECCV16} --- uses 3D U-Net type feature extractors and a differentiable weighted Procrustes approach. 

For all these DNN-based methods, the question of the generalizability to point sets with arbitrary 
and different point cloud characteristics --- such as point set density and 
volumetric sampling (\textit{e.g.,} in the case of a volumetric 3D scan of a human brain, in contrast to point sets representing surfaces) --- remains open. 
We assume that no training data is available and primarily (except for Sec.~\ref{subsec:FGAvsDeepLearningMethods}) compare our technique to the methods making the same assumptions. 

\vspace{-0.2cm}
\subsection{Physics-Based Approaches}\label{ssec:physics_based_methods} 
Physics-based methods are also popular for many computer vision and graphics tasks. These methods interpret inputs as physical quantities and transform the data according to physical laws \cite{Wright1977, Sun2007, Aubry2011}. Several algorithms from different domains of computational science use the law of universal gravitation \cite{Wright1977, Kundu1999, Sun2007, Nixon2009, LopezMolina2010, MarcoDetchart2017, Zhang2019}. 
 
A modification of Wright's gravitational clustering~\cite{Wright1977} algorithm was used for image segmentation \cite{LaiYung1998}. The method of Sun~\textit{et al.}~for edge detection~\cite{Sun2007} --- later improved by Lopez-Molina~\textit{et al.}~\cite{LopezMolina2010} --- was shown to be more robust in scenarios with noise, compared to several competing methods. 
A gravitational analogy was also applied in image smoothing \cite{MarcoDetchart2017}.
Recently, \cite{Zhang2019} proposed a weighting scheme with a gravitational model for stereo matching. 

Jauer \textit{et al.}~\cite{Jauer2019} developed a framework for RPSR based on the laws of mechanics and thermodynamics. Similarly to our FGA, they model point clouds as rigid bodies of particles and additionally support arbitrary driving forces such as gravitational or electromagnetic, as well as repulsive forces. To reduce the runtime, they apply simulated annealing along with the Monte-Carlo re-sampling and calculate particle interactions in parallel on a GPU. This policy does not improve upon the quadratic complexity. In contrast, we reduce the computational complexity of particle interactions to quasilinear, and our parallelization on a GPU further lowers the runtime. 
At the same time, we obtain positional updates for particles by solving second-order ODEs.

BHRGA \cite{BHRGA2019} is another gravitational approach for RPSR that modifies laws of gravitational attraction and converts it to non-linear least squares (NLLS) optimization problem with globally multiply-linked point interactions \cite{BHRGA2019}. Similar to our method, they also adapt a BH tree \cite{1986Natur324446B}. Relying on a Gauss-Newton solver for NLLS has both advantages and downsides. Even though \cite{BHRGA2019} requires a smaller number of iterations until convergence on average compared to FGA, our method enables fine-grained control over the parallelization and is tailored for a single GPU. We require ${\sim}1$ second for LiDAR data alignment, whereas BHRGA needs ${\sim}1.5$ minutes~\cite{BHRGA2019} for the inputs of the same size, \textit{i.e.,} an improvement of two orders of magnitude.

In another physical interpretation, \cite{golyanik2020quantum} solves small correspondence problems on point sets using quantum annealing \cite{Farhi2001}.

\vspace{-0.2cm} \section{{\Large $\mathtt{n}$}-Body Simulations}\label{sec:N_BodySimulation} The $\mathtt{n}$-body problem is defined for a system of $n$ astrophysical particles in a state of dynamic equilibrium following Newton's law of gravitational interactions~\cite{Diacu1996}. For a two-body system, \textit{the total work done by the gravitational force of attraction $\mathbf{F}_i$ by a stationary particle (at position $\mathbf{r}_j$ with mass $m_j$) in bringing the other particle (at position $\mathbf{r}_i$ with mass $m_i$) towards it by displacing a distance of $r$ units  is defined as the Gravitational Potential Energy (GPE) $\mathbf{E}_i = -\int_{\mathbf{r}_i}^{\mathbf{r}_j} \mathbf{F}_i dr = -\int_{\mathbf{r}_i}^{\mathbf{r}_j}\frac{Gm_i m_j }{r^2}dr$}. Analogously, for the case of $n$ particles the total GPE of an $\mathtt{n}$-body system is defined as: 
\begin{equation}\label{eqn:GeneralGPE}
 \mathbf{E} = -\sum_{\substack{i,j=1\\ i\neq j}}^{n} 
\int_{\mathbf{r}_i}^{\mathbf{r}_j}\frac{Gm_i m_j}{r^2}dr. 
\end{equation}
The total gravitational force $\mathbf{F}_i$ exerted on the particle $i$ 
by the remaining $n-1$ particles can be expressed as the sum of the negative gradients of the total GPE 
$\phi(\mathbf{r}_i,t)$ and an external potential $\phi_{ext}$ ~\cite{Trenti:2008}: 
\begin{equation}\label{eq:NbodyForce}
 \mathbf{F}_i = - \underbrace{ 
		    \sum\limits_{j=1,\,j\neq i}^{n} 
		    \frac{Gm_im_j(\mathbf{r}_i - \mathbf{r}_j)}{\left(\lVert \mathbf{r}_i - \mathbf{r}_j\rVert^2 
		    + \epsilon^2\right)^{3/2}}}_{\nabla\phi(\mathbf{r}_i,\,t)} 
	      - \nabla\phi_{ext}(\mathbf{r}_i), 
\end{equation}
where $\nabla$ denotes the gradient operator and $\lVert\cdot\rVert$ denotes $\ell_2$-norm. The instantaneous system's state is defined by $n$ position ($\mathbf{r}_i$) and velocity ($\dot{\mathbf{r}}_i$)
vectors at time $t$. The force softening parameter $\epsilon$ helps to avoid degeneracy inside the interaction region, \textit{i.e.,} when $\lVert \mathbf{r}_i - \mathbf{r}_j\rVert \leq \epsilon$. Absence of $\epsilon$ also indicates collisional particle interactions. $-\nabla\phi_{ext}$ accounts for any probable external forces due to the friction or other annealing factors in the system. This friction dissipates the fraction $\eta$ of the particle momentum. After solving the second-order ODEs of motion, the acceleration 
\begin{equation}
 \ddot{\mathbf{r}}_{i} = \frac{\mathbf{F}_i}{m_i}  
\end{equation}
provides the updated velocity and displacement as single and double integrals ($\int\ddot{\mathbf{r}}_{i}(t)dt$ and $\int\int\ddot{\mathbf{r}}_{i}(t)dt$) over time, and the trajectories of $n$ particles are obtained in this phase space\footnote{the space spanned by all possible particle states}. The interactions between particles can either be \textit{collisional} or \textit{collisionless}. The total energy, kinetic and potential, of every particle is conserved for the collisionless interactions or 
redistributed when collisions occur. The rules of the energy exchange and altered kinematics in collisional interactions~\cite{2003gnbs} will force the particles to collapse and cause topological degeneracy. 
For this reason, 
collisional dynamics is beyond the scope of the RPSR problem domain. 

\vspace{-0.2cm}
\section{The Proposed Approach}\label{sec:ProposedGA_For_RPSR} 
This section describes our FGA with algorithmic steps on (i) how a constrained $\mathtt{n}$-body simulation is fitted to define the dynamics of the source object in Sec.~\ref{subsec:NotationsAndDefinition}, (ii) the use of ODEs of dynamics to obtain minimum GPE for locally-optimal alignment between the source and the target in Secs.~\ref{subsubsec:GPE_GA}--\ref{subsubsec:Rigidity_Constraing_GA}, (iii) acceleration scheme for the whole process of $\mathtt{n}$-body simulation in Secs.~\ref{subsec:AccPolicy}--\ref{subsec:BH-Tree Force} and, finally, (iv) defining boundary conditions for our energy function using mass initialization policies based on shape descriptors in Sec.~\ref{subsec:BoundaryConditions}.

\vspace{-0.25cm} 
\subsection{Notations and Assumptions}\label{subsec:NotationsAndDefinition} 
In FGA, $\mathbf{Y}_{M\times D} = (\mathbf{Y}_1 \hdots \mathbf{Y}_M)^{\mathsf{T}}$ and $\mathbf{X}_{N\times D} = (\mathbf{X}_1 \hdots \mathbf{X}_N)^{\mathsf{T}}$ are unordered sets of $D$-dimensional $M$ \textit{template} points and $N$ \textit{reference} points, where $\mathbf{Y}_{i}$ and $\mathbf{X}_{j}$ denote elements with indices $i$ and $j$ from respective point sets. The following parameters need to be set for $\mathtt{n}$-body simulation which are inherited in our FGA and explained in Sec.~\ref{sec:N_BodySimulation}: 
\begin{center}
\fbox{\parbox{0.95\linewidth}{
\begin{center}
\begin{itemize}
\item $G$ --- \textbf{gravitational constant},
\item $\epsilon$ --- near field \textbf{force softening length}, 
\item $\eta$ --- constant system \textbf{energy dissipation rate}, 
\item $\Delta t$ --- \textbf{time integration step} for the ODEs of motion,
\item $m_{\mathbf{Y}_i}$ and $m_{\mathbf{X}_j}$ --- \textbf{point masses} of $\mathbf{Y}_i$ and $\mathbf{X}_j$.
\end{itemize} 
\end{center}
}}
\end{center}

The proposed method considers the general settings for particle interactions from \cite{Golyanik2016GravitationalAF}. It takes $\mathbf{X}$ and $\mathbf{Y}$ as input sets of points and interprets them as $(M+N)$-body system, where $\mathbf{Y}$ is optimally registered to $\mathbf{X}$ by estimating the rigid transformation tuple ($\mathbf{R,t}$). For the registration purpose, several assumptions and modifications are made for our $(M+N)$-body system:  
\renewcommand{\theenumi}{\roman{enumi}}
\begin{enumerate}
  \item\label{I1} Every point represents a particle with a mass condensed in an infinitely small volume; 
  \item\label{I2} A static reference $\mathbf{X}$ induces a constant inhomogeneous gravitational force field and its points do not interact with each other;  
  \item\label{I3} Particles $\mathbf{Y}_{i}$  move in the gravitational force field induced by all $\mathbf{X_j}$ and do not affect each other;   \item\label{I4} $\mathbf{Y}$ moves rigidly, \textit{i.e.,} the transformation of the template particle system is described by the tuple $(\mathbf{R}, \mathbf{t})$;
  \item\label{I5} A collisionless $\mathtt{n}$-body simulation is performed, since the number of particles cannot be 
  changed according to the problem definition;
  \item\label{I6} Astrophysical constants (\eg $G$) are considered as algorithm parameters;
  \item\label{I7} A portion of kinetic energy is dissipated and drained from the system --- the physical system is not isolated. 
\end{enumerate}

\vspace{-0.2cm}
\subsection{Gravitational Potential Energy (GPE) Function of FGA}\label{subsubsec:GPE_GA} 
\begin{figure}[!t] 
\begin{center}
\includegraphics[width=3.5in]{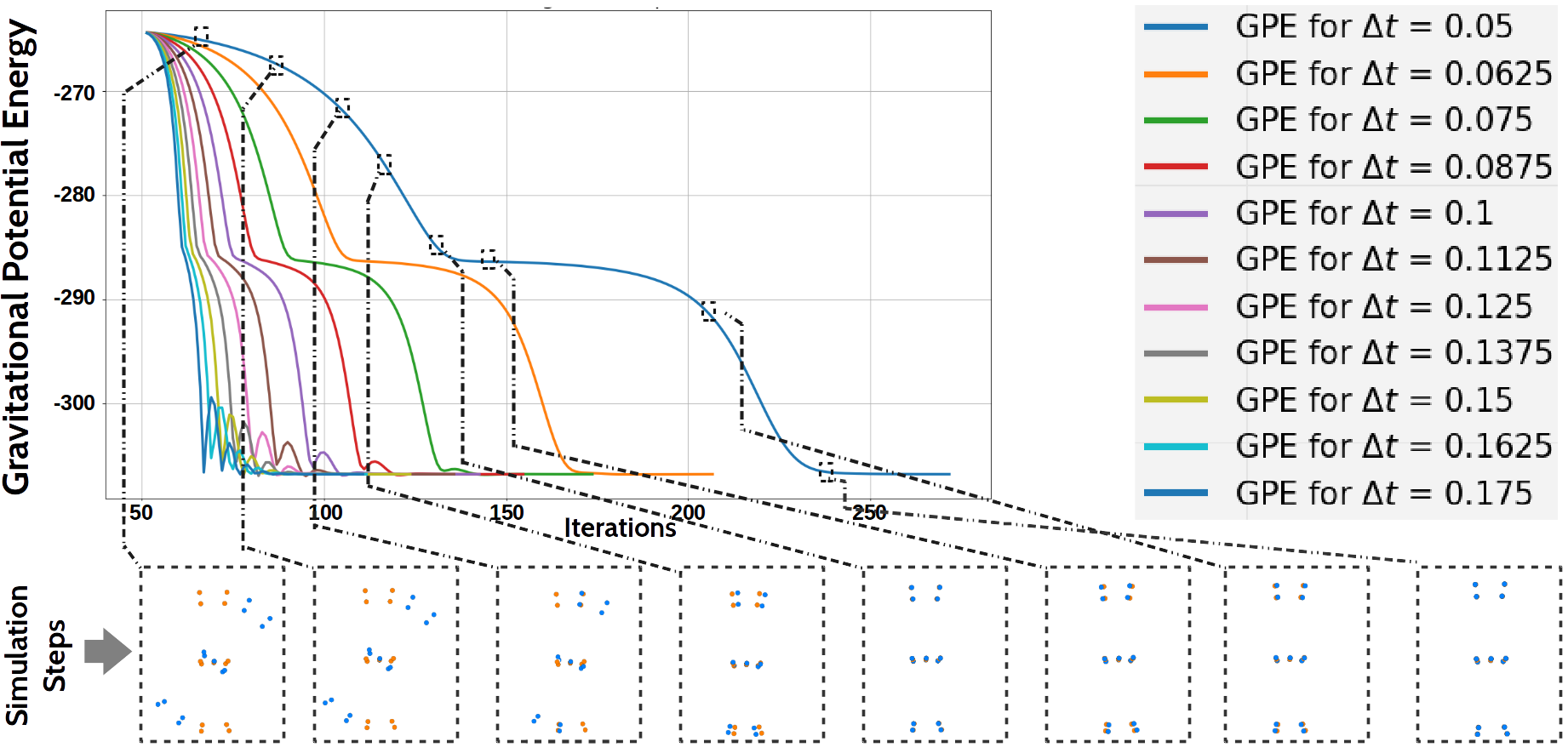} 
\end{center}
\vspace{-0.5cm}
\caption{ 
GPE function plot with different optimization \hlt{momenta} handled by the increasing time differential (time step $\Delta t$). \textit{Simulation steps} depict the reference-template configuration at the current iteration.} 
\label{fig_GPE_Time} 
\end{figure} 
The locally-optimal rigid alignment is achieved at the system's state with locally minimal GPE as shown in Fig.~\ref{fig_GPE_Time}. If the system of particles $\mathbf{Y}$ moves as a rigid body, the optimal alignment is achieved when the GPE between $\mathbf{Y}$ and $\mathbf{X}$ is minimum. We solve the rigid transformation estimation problem, \ie estimating rigid rotation $\mathbf{R}$ and translation $\mathbf{t}$, applying rigid body dynamics~\cite{ArmstrongRigidBodyDynm1985} on the displacement fields of the points in $\mathbf{Y}$. The GPE to rigidly move $\mathbf{Y}$ from its starting position towards $\mathbf{X}$ is expressed as a weighted sum of \textit{inverse multiquadric} functions on the distance fields:

\begin{equation}\label{eqn:GPE_Energy_GA}
\mathbf{E}(\mathbf{R},\mathbf{t}) = -G\,\sum\limits_{i,j}\frac{m_{\mathbf{Y}_i}\,
m_{\mathbf{X}_j}}{(\lVert\mathbf{Rr}_{\mathbf{Y}_i} + \mathbf{t} - \mathbf{r}_{\mathbf{X}_j}\rVert 
				    + \epsilon)}.
\end{equation}
Since the energy function in Eq.~\eqref{eqn:GPE_Energy_GA} is inverse multiquadric, non-linear least-squares optimization methods do not fit. Also note, there exists a singularity (without $\epsilon$) at the optimal state of $\mathbf{Y}$ as the distance field in the denominator approaches zero (whereas $\mathbf{E} \rightarrow -\infty$). We minimize \eqref{eqn:GPE_Energy_GA} using second-order ODEs of motion for particle dynamics of $\mathbf{Y}$, and iteratively recover rigid transformations on its successive states. 

\vspace{-0.2cm}
\subsection{ODEs of Particle Motion (Newtonian Dynamics)} 
The assumption (\ref{I2}) constrains $\mathbf{X}$ to remain idle at a fixed position like a single body, formed by a set of non-interactive points, which attracts $\mathbf{Y}$. On the other hand, the assumption (\ref{I3}) allows every $\mathbf{Y}_i$ to be attracted by $\mathbf{X}$ only. To this point, the dynamics of $\mathbf{Y}$ remains unconstrained, \hltt{whereas} relative positions of $\mathbf{Y}_i$ will change at every time step. The momentum of $\mathbf{Y}_i$ (from the intermediate previous state) will be preserved, and motion is tractable as the assumption (\ref{I5}) restricts any form of merging or splitting of the points in $\mathbf{Y}$. Similar to Eq.~\eqref{eq:NbodyForce}, the gravitational force of attraction exerted on $\mathbf{Y}_i$ by all the static particles from $\mathbf{X}$ reads as:

\begin{equation}\label{eq:GAForce}
 \mathbf{F}_{\mathbf{Y}_i} = -G\sum\limits_{j=1}^{N}\,
			      \frac{m_{\mathbf{X}_j} m_{\mathbf{Y}_i}(\mathbf{r}_{\mathbf{Y}_i} - \mathbf{r}_{\mathbf{X}_j})}
				    {(\lVert\mathbf{r}_{\mathbf{Y}_i} - \mathbf{r}_{\mathbf{X}_j}\rVert^2 
				    + \epsilon^2)^{3/2}}.
\end{equation}

Since gravitational force depends on the initial and final positions of a particle pair, it conserves the total mechanical energy of the system. The total energy conservation in our case is analogous to preserve the momentum of $\mathbf{Y}$ only as $\mathbf{X}$ is static. The force from $\mathbf{X}$, in Eq.~\eqref{eq:GAForce}, will bring the template $\mathbf{Y}$ closer to its centroid and then $\mathbf{Y}$ will endlessly oscillate around that centroid. Modification (\ref{I7}) allows for draining some energy in the form of heat as $\mathbf{Y}$ moves in a viscous medium. We introduce a dissipative force $\mathbf{F}_{\mathbf{Y}_i}^d$: 

\begin{equation}\label{eq:GAForceDissipative}
 \mathbf{F}_{\mathbf{Y}_i}^d = -\eta\,\dot{\mathbf{r}}_{\mathbf{Y}_i}, 
\end{equation}
which \hltt{is proportional to the particle velocity with the factor $\eta$}. It allows the otherwise endless periodic motion of the template -- due to second-order ODEs of motion -- to settle. Hence, the resultant force acting on every $\mathbf{Y}_i$ is 
\begin{equation}\label{eqn:TotalGAForce}
 \mathbf{f}_{\mathbf{Y}_i} = \mathbf{F}_{\mathbf{Y}_i} + \mathbf{F}_{\mathbf{Y}_i}^d.
\end{equation}
The next \hltt{simulated state of $\mathbf{Y}$ at the time $t+\Delta t$} in the phase space is obtained by estimating the \textit{unconstrained} velocity 
\begin{equation}\label{eqn:PreviousVelocity}
 \dot{\mathbf{r}}_{\mathbf{Y}_{i}}^{t+\Delta t}
 = \dot{\mathbf{r}}_{\mathbf{Y}_{i}}^{t} 
 + \Delta t\frac{\mathbf{f}_{\mathbf{Y}_i}}{m_\mathbf{Y_i}}, 
\end{equation}
and updating the previous $\mathbf{r}_{\mathbf{Y}_{i}}^{t}$ with the displacement 
\begin{equation}
 \mathbf{d}_{\mathbf{Y}_{i}}^{t+\Delta t} 
 = \Delta t \, \dot{\mathbf{r}}_{\mathbf{Y}_{i}}^{t+\Delta t} 
\end{equation}
for all template points. The updated velocity and displacement fields of the template points are stacked into matrices:
\begin{gather}
\mathbf{V} =  \left[ \dot{\mathbf{r}}_{\mathbf{Y}_{1}}^{t+\Delta t} \, 
		      \dot{\mathbf{r}}_{\mathbf{Y}_{2}}^{t+\Delta t} \,
		      \hdots \,
		      \dot{\mathbf{r}}_{\mathbf{Y}_{M}}^{t+\Delta t}
	      \right]^\mathsf{T} \quad \text{(velocities),}\\
\mathbf{D} =  \left[ \mathbf{d}_{\mathbf{Y}_{1}}^{t+\Delta t}  \,
		     \mathbf{d}_{\mathbf{Y}_{2}}^{t+\Delta t} \,
		     \hdots \,
		     \mathbf{d}_{\mathbf{Y}_{M}}^{t+\Delta t}
	      \right]^\mathsf{T}  \quad \text{(displacements)}
\label{eqn:CurrentVelocityAndDisplacement}
\end{gather}
and, similarly, the force residual and particle mass matrices: 
\begin{gather}
\label{eqn:forceResidualMat}
\mathbf{F} =  \left[ \mathbf{f}_{\mathbf{Y}_{1}}  \,
		     \mathbf{f}_{\mathbf{Y}_{2}} \,
		     \hdots \,
		     \mathbf{f}_{\mathbf{Y}_{M}}
	      \right]^\mathsf{T} \quad \text{(forces),}
	      \\
\mathbf{m}_{\mathbf{X}} =  \left[ m_{\mathbf{X}_{1}}  \,
		     m_{\mathbf{X}_{2}} \,
		     \hdots \,
		     m_{\mathbf{X}_{N}}
	      \right]^\mathsf{T} \quad \text{(masses of $\mathbf{X}$),}
\\
\mathbf{m}_{\mathbf{Y}} =  \left[ m_{\mathbf{Y}_{1}}  \,
		     m_{\mathbf{Y}_{2}} \,
		     \hdots \,
		     m_{\mathbf{Y}_{M}}
	      \right]^\mathsf{T} \quad \text{(masses of $\mathbf{Y}$).} 
\label{eqn:GAForceResidual_Matrix}
\end{gather}

\subsection{Rigid Body Dynamics using ODEs of Motion}\label{subsubsec:Rigidity_Constraing_GA}
Newton-Euler equations of motion in mechanics (first, second and third law of motion with Euler time integration) relate any external force $\mathbf{f}$ with the inertial state of the body. Newtonian mechanics assumes an inertial frame of reference which is fixed and excluded from any external force. According to assumption (\ref{I2}), the template and the reference are attached to the moving \textit{body-fixed} and the inertial frames of reference, respectively. After one step of our $(M+N)$-body simulation, two states of the template are available --- the previous state $\mathbf{Y}$ and the current state $\mathbf{Y} + \mathbf{D}$. To recover a single consensus rigid transformation between these two states, we solve the \textit{generalized orthogonal Procrustes alignment} problem in the closed form. 

\vspace{3pt}
\noindent\textbf{Rotation Estimation.} Given are point matrices $\mathbf{Y}$ and $\mathbf{Y}_{\mathbf{D}} = \mathbf{Y} + \mathbf{D}$. Let $\mu_{\mathbf{Y}}$ and $\mu_{\mathbf{Y}\mathbf{D}}$ be the mean vectors of $\mathbf{Y}$ and $\mathbf{Y}_{\mathbf{D}}$ respectively, let $\hat{\mathbf{Y}} = \mathbf{Y} - \mathbf{1} \mu_{\mathbf{Y}}^T$ and $\hat{\mathbf{Y}}_{\mathbf{D}} = \mathbf{Y}_{\mathbf{D}} - \mathbf{1} \mu_{\mathbf{Y}\mathbf{D}}^T$ be point matrices centered at the origin of the coordinate system and let $\mathbf{C} = \hat{\mathbf{Y}}_\mathbf{D}^T \hat{\mathbf{Y}}$ be a covariance matrix. Let $\mathbf{U}\mathbf{S}\hat{\mathbf{U}}^T$ be singular value decomposition of $\mathbf{C}$. Then the optimal rotation matrix $\mathbf{R}$ is given by~\cite{Kabsch1976,Horn:88}: 
\begin{align}\label{eq:R_Kabsch}
  \mathbf{R} = \mathbf{U}\Sigma \hat{\mathbf{U}}^T, \, \text{where} \, \Sigma = 
  \operatorname{diag}(1, \hdots, \operatorname{sgn}(|\mathbf{U} \hat{\mathbf{U}}^T|)).
\end{align}

\noindent\textbf{Translation Estimation.} 
Once the rotation $\mathbf{R}$ is resolved, such that $\mathbf{R}\hat{\mathbf{Y}} = \hat{\mathbf{Y}}_{\mathbf{D}}$, the current state of template point set can be updated as $\mathbf{Y}_{D} = \mathbf{R}\mathbf{Y} + (\mathbf{1}\mu_{\mathbf{Y}\mathbf{D}}^T - \mathbf{R}\mathbf{1}\mu^T_{\mathbf{Y}})$ where the translation component is: 
\begin{equation}\label{eqn:transEst}
 \mathbf{t} = \mathbf{1}\mu_{\mathbf{Y}\mathbf{D}}^T - \mathbf{R}\mathbf{1}\mu^T_{\mathbf{Y}}. 
\end{equation}

\vspace{-0.2cm}
\subsection{Acceleration Policies}\label{subsec:AccPolicy} 
Many acceleration policies can be used for $\mathtt{n}$-body problems. Some studies show specialized hardware configurations for massively parallel $\mathtt{n}$-body simulations. GRAPE-4 (GRAvity PipE) and GRAPE-6~\cite{Makino1998ScientificSW} use pipelining of instructions in force computation and position update of particle set sizes up to $10^4$ and $10^6$, respectively. Logical parallelism of $\mathtt{n}$-body simulation is also achieved on FPGA~\cite{FPGA_Nbody}. Few seminal works have reduced the algorithmic complexity of the $\mathtt{n}$-body problem (\eg Fast Multi-pole Method (FMM)~\cite{GREENGARD1987325}). This runs the $\mathtt{n}$-body algorithm in $\mathcal{O}(N\log{N})$ time, but can also achieve $\mathcal{O}(N)$ at the expense of higher force approximation tolerance. It also has a relatively lower force approximation accuracy than BH method~\cite{1986Natur324446B}. Although the runtime of FMM is similar to the BH method, we adapt the BH method because its underlying concept is simple and many astrophysical particle simulations using BH method are also successfully ported on FPGA and GPU~\cite{spurzem2009accelerating, WINKEL2012880}. The algorithmic steps of our RPSR method are not exactly the same as $\mathtt{n}$-body algorithm. Hence, in this section, we show how BH tree can be applied to FGA. We are the first, to the best of our knowledge, to decompose the quadratic computational complexity of $\mathcal{O}(MN)$ for all-to-all force computations to the quasilinear complexity of $\mathcal{O}(M\log{N})$ using the BH force approximation in a gravitational point set alignment method with second-order ODEs. 

\vspace{-0.2cm}
\subsection{Barnes-Hut Force Approximation}\label{subsec:BH-Tree Force} 
\begin{figure}[!t] 
\begin{center}
   \includegraphics[width=0.90\linewidth]{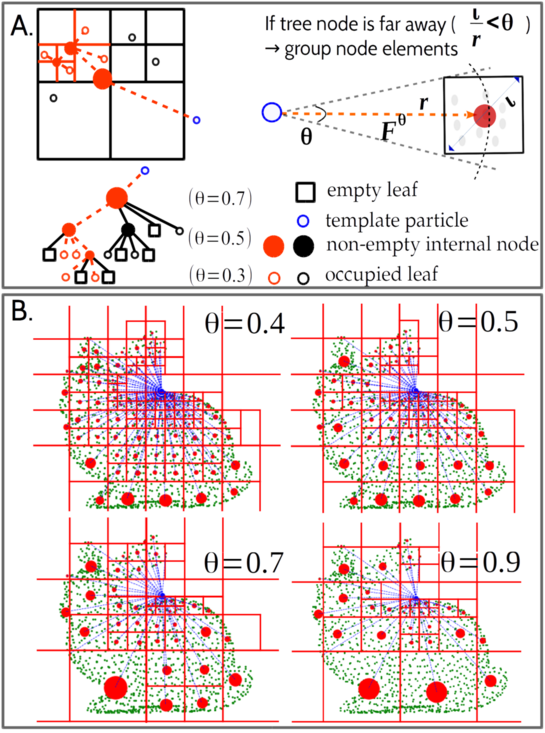}
\end{center}
\vspace{-0.5cm}
\caption{(A): BH tree structure with hierarchical cell grouping and pruning depends on the
   node opening criteria $\theta$. (B): BH force approximation $\mathbf{F}^\theta$~\cite{1986Natur324446B} under 
   different values of the node opening parameter $\theta$ for a \textit{template} point set (in green color). The 
   red dots represent the CoM of the cells with sizes proportional to the number of points inside the nodes.} 
\label{fig:BH_FORCE_TREE} 
\end{figure} 
BH-tree-based force computation in FGA requires two steps --- BH tree construction and BH force calculation. 

\noindent\textbf{BH Tree} $\mathbf{\tau}^\theta$. The construction of $\mathbf{\tau}^\theta$ is a method 
of hierarchical space partitioning and grouping of particles as described in Alg.~\ref{alg:BHtreebuild} \cite{1986Natur324446B}. 
Given a system of particles with their known position vectors, masses and dimensionality ($D$) as input, this 
method at first constructs a tree structure of virtual $2^D$ dimensional lattices, \ie either \textit{cubical}
cells as nodes of an octree for 3D data or \textit{square} cells as nodes of a quad-tree for 2D data. 
To construct $\mathbf{\tau}^\theta$, the following steps are performed iteratively until each particle is assigned 
to its leaf nodes: 

\begin{enumerate}
 \item BH tree partition starts from the center of the particle system as root node, which encapsulates 
 all the points. With this encapsulation, we split the cell in $2^3$ cubical (3D) 
 or $2^2$ quad (2D) sub-cells. This opens up \textit{three} options on the occupancy status --- either the 
 newly-created sub-cells are \textit{empty}, \textit{non-empty} (with multiple particles) or have exactly \textit{one particle}. 
 \vspace{0.1cm}
 \item For \textit{non-empty} cells, calculate the aggregated masses of the particles as
 the mass of the cell/node and the \textit{center of mass} (CoM) which is a weighted average of the
 particle positions and their respective masses. These two attributes and the cell's length are associated 
 with every cell. For a singleton cell, no further action is taken.
 \vspace{-0.25cm}
 \item For non-empty cells, we progress with further sub-divisions by repeating step i). 
 If duplicate points exist, the partitioning can continue up to infinite depth. To avoid this, we normalize inputs
 within a numerical range of $[a, b]$ before building the tree as per Alg.~\ref{alg:BHtreebuild},
 and recurse until tree depth of $d = 20$ during construction. This sets a floating-point precision 
 range from zero to $\frac{(b - a)}{2^d}$ on the minimum distance between two particles. The interpretation of this limit on the floating-point precision 
 is that if the distance between two particles in a cell is ${<}\frac{(b - a)}{2^d}$, we do not branch 
 further into higher depth. If we do not scale the particle system in the range $[a, b]$, an arbitrarily high 
 depth limit (\hbox{\textit{e.g,}} $d = 128\,\,\text{or}\,\,256$) can be required to stop further branching.  
 After scaling the input in the range $[a, b]$, we must obtain the values of the estimated translation parameters 
 at the original scale (see our appendix). \end{enumerate}

\noindent Consider the extreme case where the positions of points in the \textit{reference} are 
uniformly spaced along the axes forming a regular grid type pattern. Then, we can claim that all the nodes 
at any level of tree depth will be non-empty and contain the maximum possible number of particles inside. 


\begin{theorem}\label{lemma1} 
The maximum number of child nodes up to the depth level $d$, except the root and the leaves,
of the BH tree $\tau_{\theta}$ built on $\mathbf{X}_{N\times D}$ with maximum depth $d_{\circ}$ 
is equal to $\sum_{d=1}^{d_{\circ}-1}(2^D)^d$. Hence, the maximum number of possible nodes 
$N_{\circ}$ of a BH tree built on an unstructured point cloud is $N_{\circ} = (N + 
\sum_{d=1}^{d_{\circ}-1}(2^D)^d + 1)$. 
\end{theorem}

\begin{algorithm}[ht]\label{alg:BHtreebuild}
\DontPrintSemicolon
\footnotesize
 \KwIn{reference $\mathbf{X}_{N\times D}$, masses $\mathbf{M}_\mathbf{X}$, maximum tree depth $d$} 
 \KwOut{BH Tree $\mathbf{\tau}^\theta$}
 $[\mathbf{X}_{min},\mathbf{X}_{max}] \leftarrow$ compute bounding box (BB) of $\mathbf{X}$\;
 $\mathbf{\tau}^\theta \leftarrow$ \texttt{CreateChild}$(\text{root}(\mathbf{\tau}^\theta),\,N, \mathbf{X}_{min},\, \mathbf{X}_{max})$\;
\SetKwFunction{FMain}{CreateChild}
 \SetKwProg{Fn}{Function}{:}{}
   	\Fn{\FMain{node$(\mathbf{\tau}^\theta),\,N, \mathbf{X}_{min},\, \mathbf{X}_{max}$}}{
	    \If{$N > 0\, \,$ and $\,\text{depth} < 20$ }{
		   \If{$N > 1$}{
		      node partition center $\mathbf{o} = \mathbf{X}_{min} + \frac{\mathbf{X}_{max} - \, \mathbf{X}_{min}}{2}$\;
		      \For{$i=1$ \KwTo $2^D$}{
		           \hbox{\footnotesize $[\mathbf{X}_{min}^i,\,\mathbf{X}_{max}^i]\leftarrow$ BB using $\mathbf{X}_{min}, \mathbf{X}_{max}, \mathbf{o}$}\;
			   $N \leftarrow$ number of points inside $[\mathbf{X}_{min}^i,\,\mathbf{X}_{max}^i]$\;
			   \texttt{CreateChild}(node$(\mathbf{\tau}^\theta).\text{child}_{i},N,\mathbf{X}_{min}^i,\,\mathbf{X}_{max}^i)$
		      }
		   }
		    $\text{node}(\mathbf{\tau}^\theta).l\leftarrow  \lVert\mathbf{X}_{max}-\mathbf{X}_{min}\rVert_{2}$\;
		    $\text{node}(\mathbf{\tau}^\theta).\text{mass}\leftarrow \sum\limits_{j}m_{\mathbf{X}_{j}},\,  \mathbf{X}_{j}
		       \in [\mathbf{X}_{min},\mathbf{X}_{max}]$\;
		    $\text{node}(\mathbf{\tau}^\theta).\text{CoM}\leftarrow
		       \sum\limits_{j}\frac{\mathbf{X}_{j}m_{\mathbf{X}_{j}}}{\text{node}(\mathbf{\tau}^\theta).\text{mass}},
		       \, \mathbf{X}_{j} \in [\mathbf{X}_{min},\mathbf{X}_{max}]$\;		
	    }
    \textbf{return} $\mathbf{\tau}^\theta;$\;}
\caption{\bf Build BH tree}
\end{algorithm}

\noindent\textbf{BH Force} $\mathbf{F}^\theta$. 
Approximation of gravitational force between distant particles in a one-to-many fashion, Eq.~\eqref{eq:GAForce},
is computed with tree-based near-field and far-field approximation. For a given particle, we always start 
from the root node of the BH tree $\mathbf{\tau}^\theta$. Forces will now be 
compounded over the child nodes recursively. During the recursion, a child node will be searched in higher depths or not depend upon the node opening 
criteria or \textit{multi-pole acceptance criteria} $\theta$. In~\cite{1986Natur324446B}, $\theta$ is 
set as a lower bound of the ratio between the length ($l$) of a cell and the distance ($r$) between query 
point to the CoM of that cell. It indicates that particles from a node 
will be merged into one particle (see Fig.~\ref{fig:BH_FORCE_TREE}-(A)-(\textit{right}))  
\textit{without} further recursions if 
\begin{equation}\label{eqn:MAC_nodeOpeningCriteria} 
\theta > \frac{l}{r}.
\end{equation}
Thus, a non-empty and non-singleton node which satisfies this inequality, merges the encapsulated particles
inside it into one with a heavier mass located at their CoM. Hence, the sum of the \textit{gravitational force residuals}
from the encapsulated particles can now be approximated by the force from the merged one (Fig.~\ref{fig:BH_FORCE_TREE}-(B), red cells). 
This results in a dropout of the numerical accuracy of forces, but also in a runtime speed-up. 
According to the problem statement, the system is not self-gravitating. 
This implies that $\mathbf{X}$ remains static, and the state change of $\mathbf{Y}$ has no effect 
on $\mathbf{X}$. Thus in FGA, we only build $\mathbf{\tau}^\theta$ once on $\mathbf{X}$ 
and use it to approximate $\mathbf{F}_{Y_i}$ in Eq.~\eqref{eq:GAForce} as 
$\mathbf{F}^{\theta}_{Y_i}$, for all $\mathbf{Y}_i$ in every iteration. It reduces the memory complexity and the overall runtime as we do not need to build the BH tree multiple times.

\vspace{-0.1cm}
\subsection{Defining Boundary Conditions}\label{subsec:BoundaryConditions} 
If available, additional alignment cues (\textit{e.g.,} point colors \cite{Danelljan2016} and prior correspondences \cite{Golyanik2016rECPD}) 
can be embedded in several RPSR methods \cite{BeslMcKay1992, MyronenkoSong2010} as boundary conditions to guide the alignment. 
A variant of rigid ICP~\cite{Korn2014ColorSG} generalizes the Euclidean distances for the color space, 
whereas a variant of CPD defines the GMM in the color space of a point cloud~\cite{Danelljan2016}. 
Extrinsic features (\eg tracked marker locations) over multiple frames~\cite{Bagchi2019} can also increase the correspondence reliability during the registration. 
In contrast, point-based \cite{Rusu2009} or geometric features~\cite{Khoury2017LearningCG} extracted from the input can help to boost the correspondence search. 
Whereas it is seemingly useful for~\cite{Rusu2008} to add feature descriptors from~\cite{Rusu2009} for robust registration, 
Golyanik~\etal~\cite{Golyanik2016rECPD} suggest to define a separate GMM probability density function for a prior set of landmarks. 
\cite{Serafin7759604} can extract the sensor's viewpoint signatures as well as the underlying 
geometry descriptors from considerably noisy point clouds, which helps to estimate 6 DOF camera pose in~\cite{aldoma2011cad}. 

We define the boundary conditions in two ways, and show how they guide $\mathbf{Y}$ for robust alignment, especially \textit{in the case of partially overlapping and noisy data.} 
We use (1) a set of known one-to-one correspondences as landmarks 
and (2) point cloud feature-based weights as point masses. We finally derive a (3) Smooth-Particle Mass (SPM) map using RBFs and NIV meausres for assigning weights to each point as its input masses.  

\subsubsection{RBF Mass Interpolation}\label{subsec:RBFMassInterpolant}
An RBF~\cite{doi:10.1029/JB076i008p01905} is a radially symmetric function around a central point.
For any given input vector $\mathbf{Y}_i$, 
an RBF interpolation using a radial kernel \hbox{$\Phi: \mathbb{R}^D\times\mathbb{R}^D \rightarrow \mathbb{R}$} with $m$
radial centroids (suppose these are originating from prior matches -- $\mathbf{Y}_{\widehat{c}^1}\hdots\mathbf{Y}_{\widehat{c}^m}$, where \hbox{$C = \{(\widehat{c}^1, \tilde{c}^1), \hdots, (\widehat{c}^m, \tilde{c}^m)\}$} is a set of correspondence indices with entries $\widehat{c}_Y \in \left\lbrace1,\hdots, M\right\rbrace$ and $\tilde{c}_X \in \left\lbrace1,\hdots, N\right\rbrace$) is defined as:
\begin{equation}\label{eqn:RBF_function} 
 \mathbf{B}(\mathbf{Y}_i) =\begin{cases} 
                          \sum\limits_{\widehat{c}\in C}\lambda_{i\widehat{c}}\Phi(\lVert\mathbf{Y}_{i} - \mathbf{Y}_{\widehat{c}}\rVert) 
                          & \text{if $C\neq\emptyset$ }\\ 
                          1.0, & \text{if $C=\emptyset$}. 
                         \end{cases} 
\end{equation} 
RBFs are invariant under Euclidean transformations, which is suitable for embedding in
iterative alignment methods, especially for adaptive mass computation. 
In FGA, we choose $\Phi(s) = \exp\left(-\frac{s^2}{\sigma^2}\right)$. The computational complexity of evaluating $\mathbf{B}$, which is to determine $\lambda_{i\widehat{c}}$
by collocation on all $M$ \textit{template} points from a sparse set of
$m$ anchor points, is $\mathcal{O}(Mm^2)$ ($\mathcal{O}(Nm^2)$ for $\mathbf{X}$). 
Note that $m$ is usually small. 
The interpolated masses with three centroids are highlighted in Fig.~\ref{fig:SPM_ONLY}-A-(\textit{middle}). 

\subsubsection{Normalized Intrinsic Volume Measure}\label{subsec:VoxelDensityNormalization}
\begin{figure}[!t]
\begin{center}
   \includegraphics[width=0.99\linewidth]{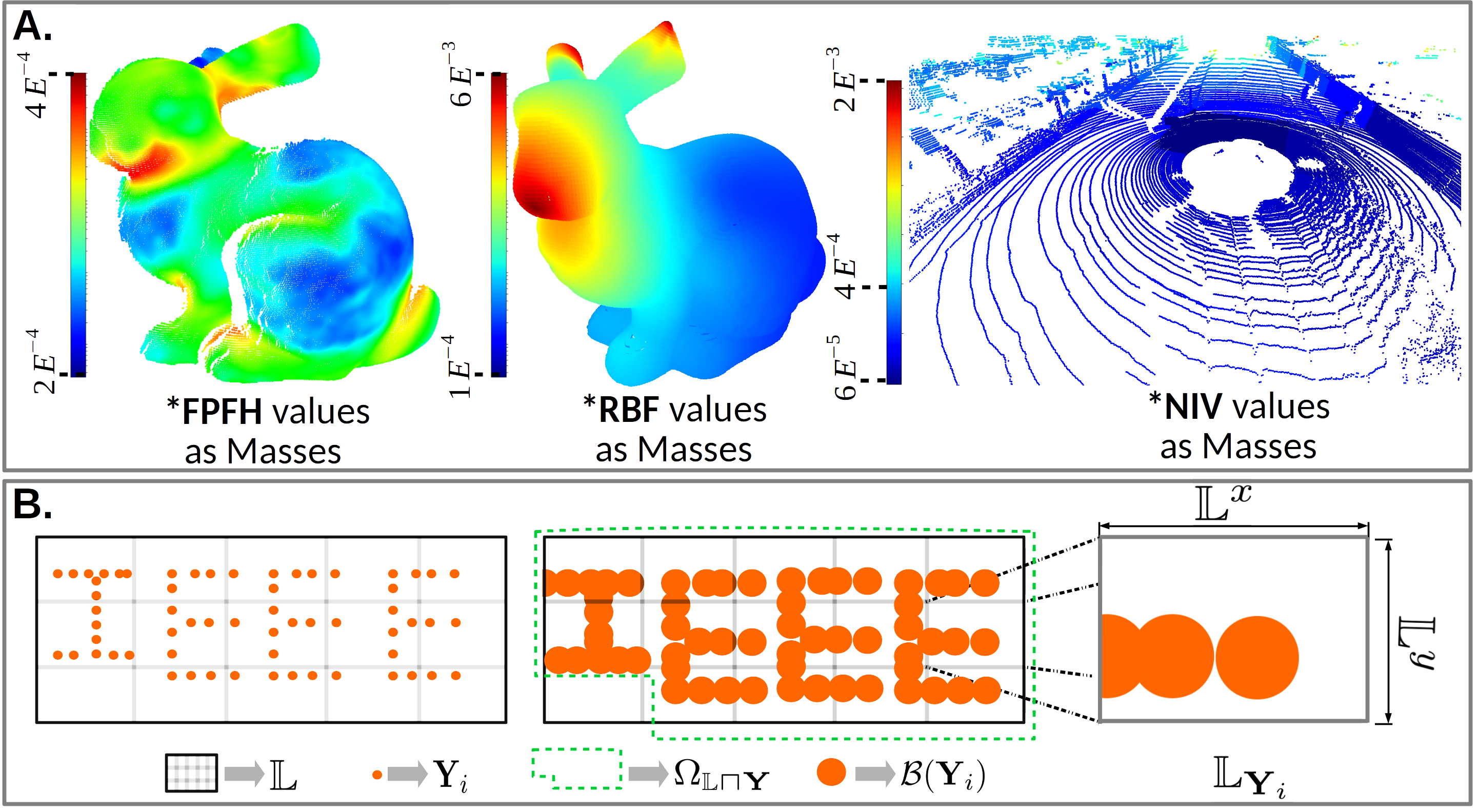}
\end{center}
   \caption{(A): The SPM function $\mathbf{S}$ maps different point-wise feature values to the masses. 
   Shown are the FPFH~\cite{Rusu2009} and   
   the estimated RBF~\cite{doi:10.1029/JB076i008p01905} values using landmarks, as well as NIV 
   measures for the mass initialization of \textit{bunny} and KITTI~\cite{Geiger2013IJRR} data. 
   (B): A lattice representation of a 2D point set  
   and the visualization of the intrinsic volume defined as the fraction    of the cell area occupied by all $\mathbf{Y}_i$-centric balls $\mathcal{B}(\mathbf{Y}_i)$ 
   inside it.} 
\label{fig:SPM_ONLY}
\end{figure}

The first step is to discretize each $D$-dimensional input point cloud into $\varrho^D$ equispaced 
lattices $\mathbb{L}$, where $\varrho = 16$ in our method. These lattices contain scattered density 
information across the domains $\Omega_{\mathbf{X}}$ and $\Omega_{\mathbf{Y}}$ of $\mathbf{X}$ and $\mathbf{Y}$, 
respectively. Each of the non-empty lattice cells in $\mathbb{L}$ that cover only the domain of input point cloud,
(\eg $\Omega_{\mathbf{Y}}$ of $\mathbf{Y}$) has equal intrinsic volume 
(or total quermassintegrals $V$~\cite{McMullen1991}), of either $\mathbb{L}^x \times \mathbb{L}^y$ in 2D or 
$\mathbb{L}^x \times \mathbb{L}^y \times \mathbb{L}^z$ in 3D. 
These lattices are $\mathbb{L}\sqcap\mathbf{Y}$\footnote{``$\sqcap$'' denotes set intersection on a discretized domain}, and their combined intrinsic volume is 
$\int_{\Omega_{\mathbb{L}\sqcap\mathbf{Y}}} V(\mathbb{L})$. 
Recall that the lattices are either sparsely or densely packed by the convex bodies $\mathcal{B}(\mathbf{Y}_i)$
centered at the locations of all $\mathbf{Y}_i$. 
Each of these bodies has constant intrinsic volume $V(\mathcal{B}(\mathbf{Y}_i)) 
= \pi(\frac{b-a}{2d\varrho})^2$ in 2D or $V(\mathcal{B}(\mathbf{Y}_i)) = \frac{4}{3}\pi(\frac{b-a}{2d\varrho})^3$ in 3D.   
The NIV value of $\mathbf{Y}_{i}$, contained inside the lattice cell $\mathbb{L}_{\mathbf{Y}_i}$, 
is expressed as the inverse of the ratio between the $V(\cup_{i\in \mathbb{L}_{\mathbf{Y}_i}}\mathcal{B}(\mathbf{Y}_i))$ and 
$V(\mathbb{L}_{\mathbf{Y}_i})$, and normalized by $\int_{\Omega_{\mathbb{L}\sqcap\mathbf{Y}}} V(\mathbb{L})$: 
\begin{equation}\label{eq:Normalized_IntrinsicVol}
 \mathbf{N}(\mathbf{Y}_{i}) = \left(
			      \frac{1}{{\int_{\Omega_{\mathbb{L}\sqcap\mathbf{Y}}} V(\mathbb{L})}}
			      \frac{V(\cup_{i\in \mathbb{L}_{\mathbf{Y}_i}}\mathcal{B}(\mathbf{Y}_i))}{V(\mathbb{L}_{\mathbf{Y}_i})}
			      \right)^{-1}.
\end{equation}
The area (for 2D) or volume (for 3D) fraction of 
lattices covered either by the circles (convex bodies for 2D) or spheres (convex bodies for 3D) 
centered around the enclosed points as the structure descriptors\footnote{
The motivation of this descriptor comes from the Monte-Carlo Importance Sampling technique --- suppose $m$ 
samples are drawn using some function $f$ from an alternative population (presented by probability density 
function $q(x)$) of the actual population (presented by probability density function $p(x)$), then the 
expected value of the samples $\int f(x)p(x)dx$ can also be expressed as $\int \frac{f(x)p(x)}{q(x)}q(x)dx = 
\mathbb{E}\left[\frac{f(x)p(x)}{q(x)}\right]$. The probability fraction $w(x) = \frac{p(x)}{q(x)}$ 
are the importance weights, such that the normalized importance is $\sum_{i=1}^m w(x_i) = 1$.} is termed as 
\textit{normalized intrinsic volume} (NIV) measure.  
Fig.~\ref{fig:SPM_ONLY}-A-(\textit{right}) shows color-coded NIV values of LiDAR scan points 
and the NIV measure details are depicted in Fig.~\ref{fig:SPM_ONLY}-B.

\subsubsection{SPM: Smooth-Particle Mass Function}\label{subsec:Smooth-Particle Mass} 
FGA assigns point-wise features as their masses. If landmarks are available, 
we choose strong weights on them and radially interpolated weights on the rest of the points using the RBF 
map~\cite{doi:10.1029/JB076i008p01905} $\mathbf{B}(\mathbf{Y}, C): \mathbb{R}^{M\times D} \rightarrow 
\mathbb{R}^{M\times1}$. If no landmarks are available, we use NIV measure
$\mathbf{N}(\mathbf{Y}): \mathbb{R}^{M\times D} \rightarrow \mathbb{R}^{M\times1}$. 
The values of the feature maps are spatially smooth. In some scenarios when point densities are 
non-uniform and prior matches are available, then the Hadamard product, symbolized by $\circ$, of RBF and NIV 
\begin{equation}
\label{eqn:FinalSPM}
 \mathbf{S} = \mathbf{N}\circ
  \mathbf{B}
 \end{equation}
assigns balanced weights on $\mathbf{X}$ and $\mathbf{Y}$. Alternatively, we can also map other 
point-based or geometry-based feature values (\eg~FPFH~\cite{Rusu2009} or CGF~\cite{Khoury2017LearningCG}) as 
SPMs to the masses, which updates the matrices $\mathbf{M}_{\mathbf{X}} = \mathbf{S}(X)$ and 
$\mathbf{M}_{\mathbf{Y}} = \mathbf{S}(Y)$. FGA is summarized in Alg.~\ref{alg:FGA_Algorithm}.

\SetKwInOut{Parameters}{Parameters}
\begin{algorithm}
\DontPrintSemicolon
\footnotesize
\KwIn{reference $\mathbf{X}_{N\times D}$, template $\mathbf{Y}_{M\times D}$, landmarks $C = (\widehat{c}, \tilde{c})$}
\KwOut{optimal rigid transformation $\mathbf{T}^*$ registering $\mathbf{Y}$ to $\mathbf{X}$}
\Parameters{$\epsilon$, $\eta$, $G$, $\varrho$, $\Delta t$, $\theta$, $\sigma$}
 \textbf{Initialization:} $\mathbf{T} = [\mathbf{R}=\mathbf{I}_{3\times 3}|\mathbf{t} = \mathbf{0}_{3\times 1}]$\;
 normalize $\mathbf{X,Y}$ in the range $[a,b]$\; 
 compute $\mathbf{B}(\mathbf{Y})$, $\mathbf{B}(\mathbf{X})$ using Eq.~\eqref{eqn:RBF_function}\;
 compute $\mathbf{N}(\mathbf{Y})$, $\mathbf{N}(\mathbf{X})$ using Eq.~\eqref{eq:Normalized_IntrinsicVol}\;
 $\mathbf{\tau}^\theta$ $\leftarrow$ build BH tree on $\mathbf{X}$ with SPM $\mathbf{S}(\mathbf{X})$ using Alg.~\ref{alg:BHtreebuild}\;
 \While{$
 \lVert\mathbf{T}^{t+\Delta t} - \mathbf{T}^{t-\Delta t}\rVert_{\text{F}}^2
\leq 10^{-4}$}{
   compute $\mathbf{F}^{t}$ using Eqs.~\eqref{eq:GAForce},~\eqref{eqn:TotalGAForce}, and~\eqref{eqn:forceResidualMat}\;
   $\triangleright$
     \For{ $k=1$ \KwTo $M$}
	{	    
    $\mathbf{F}^{\theta}_{Y_k}\leftarrow$ \texttt{BHForce}$(\text{root}(\mathbf{\tau}^\theta), \mathbf{Y}_{k}, f=0)$\;
    }
    compute
            $\mathbf{V}^{t+\Delta t}$ and
            $\mathbf{D}^{t+\Delta t}$
	    using Eqs.~\eqref{eqn:PreviousVelocity}--\eqref{eqn:CurrentVelocityAndDisplacement}\;
    compute transformation
	$\mathbf{T}^{t+\Delta t}\leftarrow\left[\mathbf{R}^{t+\Delta t}|\mathbf{t}^{t+\Delta t}\right]$\;
    $\triangleright$ solve
	for $\mathbf{R}^{t+\Delta t}$ using ~\cite{Horn:88, Kabsch1976, Golyanik2016GravitationalAF}, and Eq.~\eqref{eq:R_Kabsch}\;
    $\triangleright$ solve for $\mathbf{t}^{t+\Delta t}$ using Eq.~\eqref{eqn:transEst}\;
    $\triangleright$ update $\mathbf{Y}^{t+\Delta t}\leftarrow(\mathbf{R}^{t+\Delta t})\mathbf{Y}^{t} + \mathbf{t}^{t+\Delta t}$\;
    $\triangleright$ $\mathbf{R}\leftarrow(\mathbf{R}^{t+\Delta t})\mathbf{R}$; ~$\mathbf{t}\leftarrow\mathbf{t}^{t+\Delta t} + (\mathbf{R}^{t+\Delta t})\mathbf{t};$\;
  }
 $\mathbf{T}^* = \left[\mathbf{R} |\mathbf{t}\right]$\;

  \SetKwFunction{Force}{BHForce}
    \SetKwProg{Fn}{Function}{:}{}
    \Fn{\Force{node$(\mathbf{\tau}^\theta), \mathbf{Y}_{k}, f$}}{
		\eIf{$\frac{\text{node}(\mathbf{\tau}^\theta).\text{l}}
		   {\lVert\mathbf{Y}_{k} - \text{node}(\mathbf{\tau}^\theta).\text{CoM}\rVert} < \theta$ and
	    $\text{node}(\tau^\theta)$ not empty}
	{
	  $f \leftarrow f +\,$ Force between $\mathbf{Y}_{k}$ and $node(\mathbf{\tau}^\theta)$ using Eq.~\eqref{eq:GAForce}
	}
	{
	  \For{$i=1$ \KwTo $2^D$}{
	    $f \leftarrow f + $ \texttt{BHForce}$(node(\mathbf{\tau}^\theta).\text{child}_{i}, \mathbf{Y}_{k}, f)$\;
	  }
	}
	\textbf{return} $f;$
    }
  \caption{{\bf Fast Gravitational Approach}}
  \label{alg:FGA_Algorithm}
  \end{algorithm}

\vspace{-0.2cm}
\subsection{Implementation Details} 
\textit{Many-body} simulation is computationally expensive and demands scalability of its underlying 
interaction algorithm. 
Fitting BH algorithm on parallel hardware~\cite{WINKEL2012880} is not straightforward as irregular tree data
structure cannot be \hlt{mapped} easily to modern \textit{uniform memory access} architecture. We modify the 
C++/CUDA implementation blueprint for BH-tree-based $\mathtt{n}$-body simulation~\cite{BURTSCHER201175} method to use in our FGA. 
We describe only the \textit{force approximation kernel} of~\cite{BURTSCHER201175} which is changed for the CUDA/C++ 
version of FGA. The rest of the kernels --- \textit{bounding box kernel}, 
\textit{tree building kernel}, \textit{node-summarizing kernel}, \textit{node-sorting kernel} and 
\textit{rigid alignment kernel} --- for FGA are straightforward operations (\textit{i.e.,} either matrix multiplications 
or sorting).

Following Lemma~\ref{lemma1}, we allocate global memory blocks for necessary book-keeping tasks 
on $2^D N_{\circ}$ children indices, $N_{\circ}$ binary flags for empty or non-empty nodes, 
$N_{\circ}$ starting positions and $N_{\circ}$ ending positions of nodes of \texttt{integer} 
type variables. 
In total, $M$ threads are launched to compute the forces on every template point. 
In this process, no synchronization barriers are required to summarize forces.

\vspace{-0.2cm}
\section{Experimental Results}\label{sec:Experiments} 
\hl{We evaluate FGA on a wide range of real and synthetic datasets with 
multiple types of data disturbances (\textit{e.g.,} noise and partial overlaps). 
The experiments cover indoor as well as outdoor scenarios, and we compare FGA to several
state-of-the-art approaches from different method classes. 
}

\vspace{0.15cm} 
\noindent\textbf{Experimental Datasets.} 
\hl{We use the Synthetic \textit{bunny} dataset\footnote{\url{www.graphics.stanford.edu/data/3Dscanrep/}} 
from the Stanford scans repository in the runtime experiments (the version with ${\sim}35k$ 
points, see Sec.~\ref{subsec:Speed_AccuracyAnalysis}), as well as for emulating effects of 
added synthetic noises and other data disturbances (the version with ${\sim}1.8k$ points, 
see Sec.~\ref{subsec:Robustness_Against_Data_Disubances}).} 
ModelNet40~\cite{ModelNet40} is another synthetic dataset comprised of multiple instances 
of $40$ different object categories (\textit{e.g.,} \textit{plant, vase, toilet, and table}). 
It has ${\sim}9.8k$ training samples and ${\sim}2.4k$ testing samples to be used by deep learning methods. 
In Sec.~\ref{subsec:FGAvsDeepLearningMethods}, we first compare deep learning methods and our FGA (a non-neural method) 
for the generalizability across different input data. 
Stanford \textit{lounge}\footnote{\url{www.qianyi.info/scenedata.html}} 
~\cite{Choi_2015_CVPR} and \textit{Freiburg}\footnote{\url{https://vision.in.tum.de/data/datasets/rgbd-dataset}} 
~\cite{sturm12iros} datasets contain partial scans of indoor scenes generated \hl{using RGB-D sensors}. 
\hl{They are widely-used in evaluations of methods for simultaneous localization and mapping (SLAM)} 
where \hl{the ratio between the area of intersection and the whole inputs --- further referred to
as the ratio of overlaps ---} for the consecutive frames \hl{is moderate (\textit{i.e.,} $>50\%$, 
and often $>80\%$).} 
We apply FGA on these two datasets 
and another more challenging dataset of similar type with indoor scenes 
3DMatch~\cite{zeng20163dmatch} (in Sec.~\ref{subsec:3DMatch}), where 
the ratio of overlaps between the source and target frames \hl{is} lower 
(\textit{i.e.,} $<50\%$, and often $<30\%$) \hl{which makes it highly challenging 
for point set alignment methods.} 
Next, we \hl{use} several driving sequences from the 
\textit{KITTI}\footnote{\url{www.cvlibs.net/datasets/kitti/raw\_data.php}}~\cite{Geiger2013IJRR} and 
\textit{Ford}\footnote{\url{http://robots.engin.umich.edu/SoftwareData/Ford}}~\cite{Pandey:2011:FCV:2049736.2049742}       
datasets which contain non-uniformly sampled point clouds generated 
using Velodyne LiDAR sensors.

\vspace{0.15cm} 
\noindent{\bf Evaluation Criteria.} Bunny dataset provides ground-truth correspondences, and the \textit{lounge}  
dataset~\cite{Choi_2015_CVPR} provides ground-truth transformations ($\mathbf{R}_{gt}, \mathbf{t}_{gt}$). 
We \hl{calculate} the root-mean-squared error (RMSE) on the distances between registered source 
and target point clouds with known correspondences 
and angular deviation $\varphi$ for the \textit{lounge} dataset, which can be measured as the 
chordal distance between the estimated ($\mathbf{R}^{*}$) and ground-truth ($\mathbf{R}_{gt}^\mathsf{T}$) rotations or as an Euler 
angular deviation~\cite{Hartley2013}: 
\begin{equation}\label{eqn:GroundTruthAngularDev} 
 \varphi
 =
 \frac{180^\circ}{\pi}\left(\cos^{-1}\left(0.5(\operatorname{trace}(\mathbf{R}_{gt}^\mathsf{T} 
 \mathbf{R}^{*}) - 1)\right)\right). 
\end{equation} 
For KITTI~\cite{Geiger2013IJRR} and 3DMatch~\cite{zeng20163dmatch} datasets, 
we measure the angular deviation $\varphi$ and Euclidean distance $\mathbf{\Delta t}$ between the 
translation components $\mathbf{t}^{*}$ (estimated) and $\mathbf{t}_{gt}$ (ground truth). 
We also report the total transformation error: 
\begin{equation}\label{eqn:Total_TransError}
\Delta\mathbf{T}
  =
 \varphi
 + 
 \underbrace{\lVert\mathbf{t}_{gt} - \mathbf{t}^{*}\rVert}_{\mathbf{\Delta t}}
\end{equation}
in the experiments for parameter selection in Sec.~\ref{subsec:OptimalParamChoices}, 
where the angular error $\varphi$ is a small residual part of $\Delta\mathbf{T}$. 

\vspace{0.2cm} 
\noindent\textbf{Baseline Methods and Parameter Settings.} 
\hl{FGA is a \textit{general-purpose} registration method which does not require training 
data and which performs equally well on volumetric point clouds and also on data with well-defined \hl{surface} geometry.} 
\hl{We hence focus on comparisons to methods making the same assumptions and report 
alignment results of CPD~\cite{MyronenkoSong2010}, 
GMMReg~\cite{GMMReg1544863}, FilterReg~\cite{Gao2019}, FGR~\cite{FGRECCV16}, 
RANSAC~\cite{Rusu2009}, point-to-point ICP~\cite{BeslMcKay1992} and 
GA~\cite{Golyanik2016GravitationalAF} as well as our FGA.} 
All these methods, like our FGA, do not require information on prior correspondences and any 
special geometric knowledge about the inputs. 
\hl{However, if available, prior correspondences can be used by FGA as boundary conditions.} 
\hl{ 
We also include a few comparisons on ModelNet40~\cite{ModelNet40} dataset where deep-learning-based
methods such as PointNetLK~\cite{AokiGSL19} and Deep Closest Point (DCP)~\cite{wang2019deep} 
are exposed to generalization gaps, \textit{i.e.,} they perform poorly when tested on data 
with point disturbances not observed in the training data. 
We show that such neural methods make strong assumptions about the types of data they can deal with. 
Hence, they are not in the focus of this article. 
}

FGA runs on \hl{a heterogeneous platform} with CPU (C++ code) and GPU (in CUDA/C++ code). 
All experiments are performed on a system with the Intel Xeon E3-1200 CPU with 16GB RAM and 
NVIDIA 1080 Ti graphics card. 
We scale our input data in the range $[-5,5]$ to build the BH tree on the reference $\mathbf{X}$ 
up to a fixed depth of $d=20$, as mentioned in Sec.~\ref{subsec:BH-Tree Force}. 
\hl{In all experiments}, RBF kernel \hl{width} ($\sigma$), NIV lattice resolution ($\varrho$), 
gravitational constant ($G$), gravitational force softening length ($\epsilon$), time 
integration step ($\Delta t$), force damping constant ($\eta$) and BH-cell opening criteria 
($\theta$) are set as --- $\sigma = 0.03$, $\varrho=16$, $G=66.7$, $\epsilon= 0.2$, 
$\Delta t = 0.1$, $\eta = 0.2$ and $\theta = 0.6$. 
Finally, we show through experiments with different datasets that these settings are optimal. 
FGA converges in $80-100$ iterations. 

\vspace{-0.15cm}
\subsection{FGA against Deep Learning Methods} 
\label{subsec:FGAvsDeepLearningMethods}
\begin{figure}[!h] 
 \begin{center} 
   \includegraphics[width=0.90\linewidth]{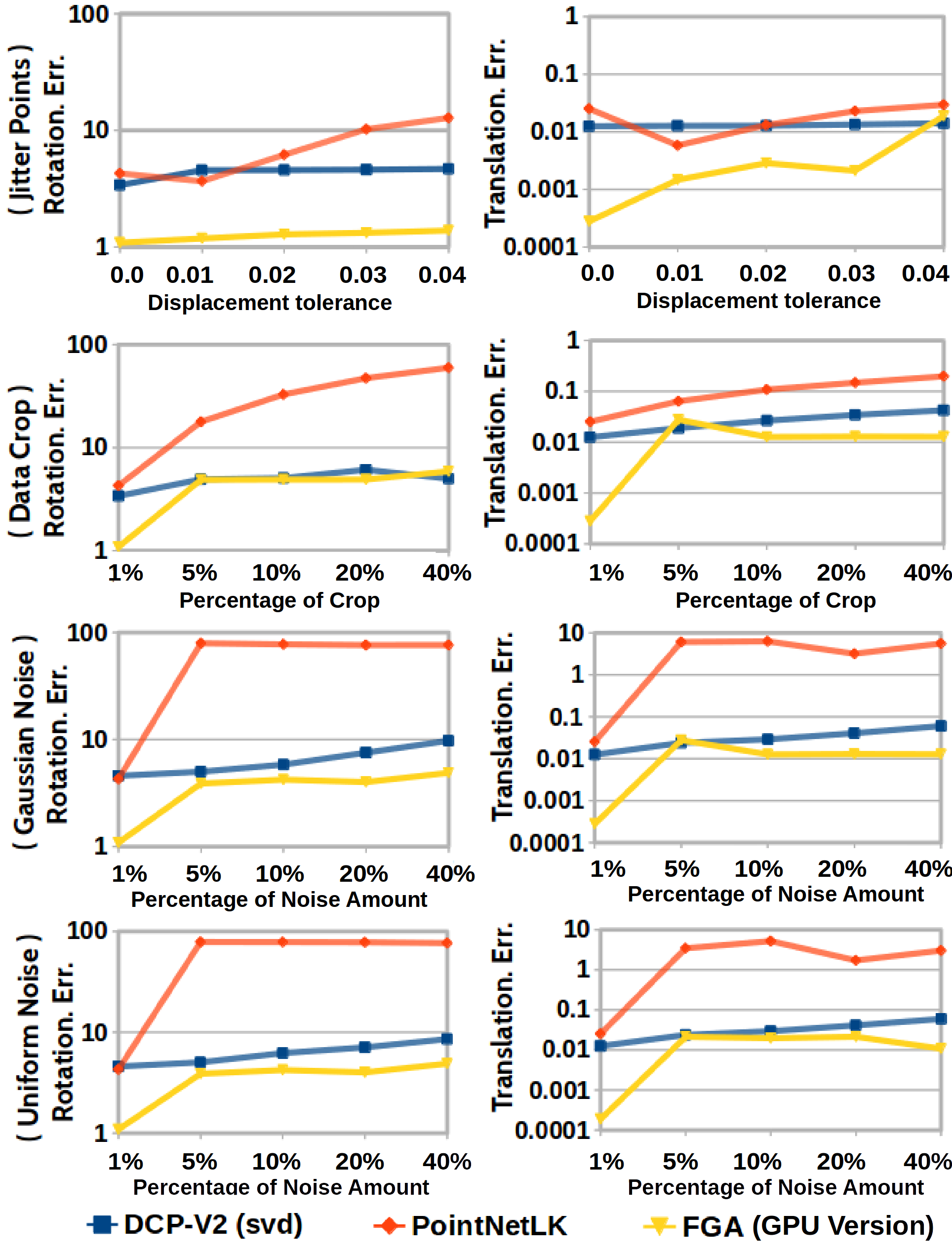} 
  \end{center} 
  \caption{The accuracy of our FGA and two deep learning methods PointNetLK~\cite{AokiGSL19} 
  and DCP~\cite{wang2019deep} trained on ModelNet40~\cite{ModelNet40} dataset, 
  with additional $10\%$ of the samples included in \hltt{the} training set after applying 
  \textit{four} different types of data disturbances. 
      Our FGA is compared on \textit{five} different validation sets with increasing levels of data disturbances. 
    The error plots show that FGA outperforms the other two methods and highlight 
  the robustness issues of the learning-based approaches.} 
\label{fig:m40_EvalDeepLR} 
\end{figure} 

This section provides a detailed analysis of the registration results using state-of-the-art 
deep learning methods --- PointNetLK~\cite{AokiGSL19} and DCP~\cite{wang2019deep} on 
the ModelNet40~\cite{ModelNet40} dataset. 
Both methods are trained from scratch (to account for noisy samples) 
for $250$ epochs with a learning rate of $10^{-3}$ using ADAM optimizer \cite{Kingma2015}. 
While a batch size $32$ is used for PointNetLK with ten internal iterations, 
DCP is trained with batch size $10$ (as recommended in \cite{AokiGSL19,wang2019deep}). 
Scalability of processing large point clouds is a common problem for both these methods 
(also, both networks have to use a fixed number of multi-layer perceptrons (MLPs) to match 
the embedding dimensions). 
Hence, we subsample all CAD shapes to $2048$ points. To enhance the accuracy of both the networks in handling data with disturbing effects, 
we perform data augmentation. 
We randomly select $950$ CAD objects from the training set of ${\sim}9.8k$ samples 
and merge them into a single extended training set 
after applying four different types of data disturbances --- (i) adding $10\%$ Gaussian 
noise with zero mean and the standard deviation of $0.02$, (ii) adding $10\%$ (out of $2048$ points)
uniformly distributed noise in the range $[-0.5, 0.5]$, (iii) adding perturbations 
to the actual point positions with maximum displacement tolerance of $0.01$, and 
finally (iv) removing $20\%$ of the points in a chunk at random. 
The choice of applying the above four disturbances on any given sample is random 
(in all of the above cases, the \hltt{total amount} of points remains $2048$ for each CAD sample). 
For comparing the errors, we prepare five different validation sets --- originating 
from the same test set ---  with increasing levels of the aforementioned 
data disturbance types --- \textit{i.e.,} by adding 
$1\%, 5\%, 10\%, 20\%,$ and $40\%$ noisy points or cropping the same amount of points.

In spite of training with the additional $950$ samples, both DCP and PointNetLK show a 
common generalizability issue, see the transformation error plots in Fig.~\ref{fig:m40_EvalDeepLR}. 
The evaluation shows that error metrics of PointNetLK becomes significantly 
higher when the noise level increases just from $1\%$ to $5\%$ (for Gaussian noise, 
the rotational error increases from $4.37^\circ$ to $79.3^\circ$, and the translational 
error increases from $0.0252$ to $6.05$, whereas for 
uniform noise, the rotational error increases from $4.29^\circ$ to $78.42^\circ$, and 
the translational error increases from $0.0249$ to $3.43$). 
While DCP approach is more robust than PointNetLK, the noise intolerance issue is still pertinent for it. 
Our FGA is far more robust compared to both neural approaches, \textit{i.e.,} 
its transformation estimation errors, especially rotational errors, are consistently 
smaller by more than $15$ times compared to PointNetLK and ${\approx}1.2$ times 
compared to DCP. 
Only the rotational error of FGA is close --- but still lower at several increasing \hltt{noise} 
levels --- compared to the DCP's error for cropped inputs.

We conclude that there exist generalization gaps and robustness issues in these 
neural approaches when tackling noisy data which is often encountered in practical applications. 
Hence, further in the experimental section, we keep our focus on unsupervised and 
general-purpose RPSR methods which do not require training data and can generalize 
across various alignment scenarios. 

\vspace{-0.2cm}
\subsection{FGA Runtime and Accuracy Analysis}\label{subsec:Speed_AccuracyAnalysis} 
\begin{figure}[!ht]
\begin{center}
   \includegraphics[width=0.92\linewidth]{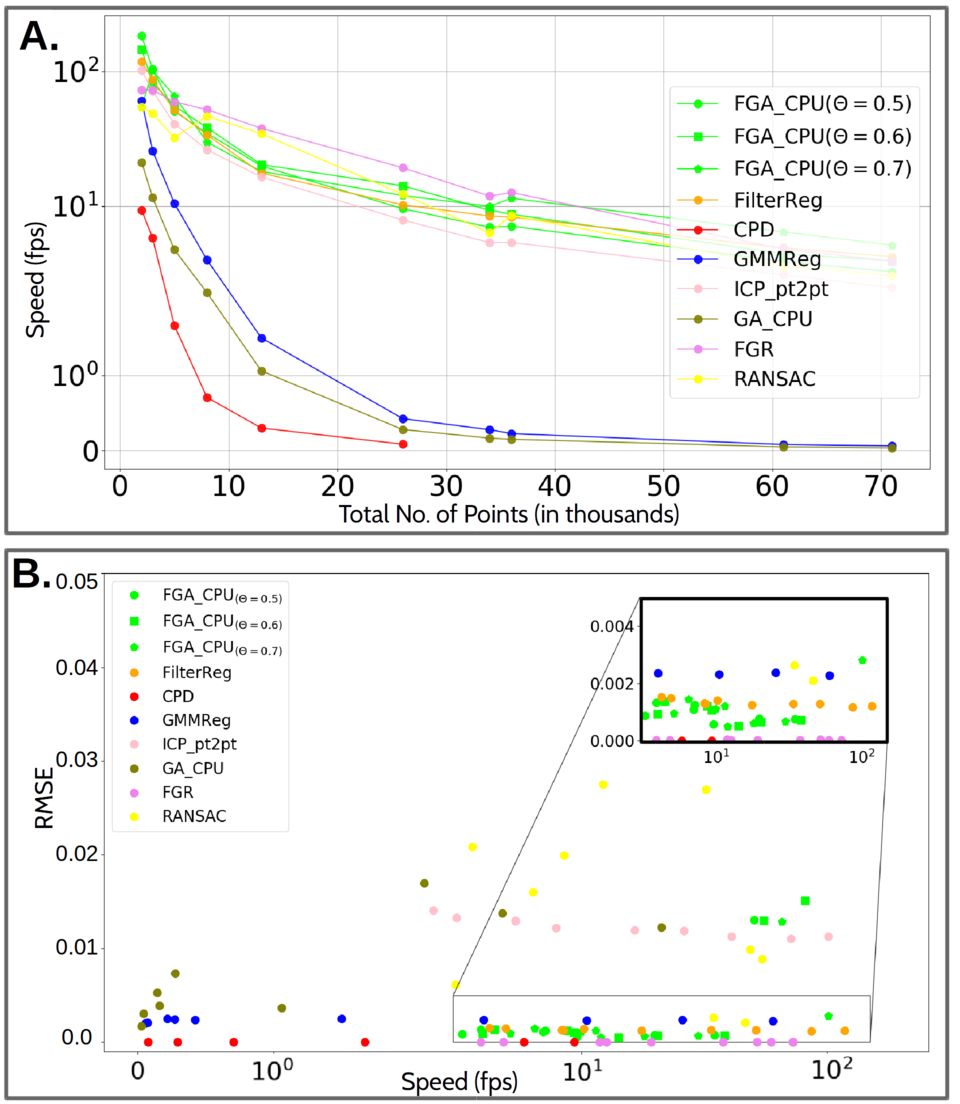}
\end{center}
\caption{(A:) Throughput of different methods in frames per second and (B:) 
Bivariate correlation plot for the alignment accuracy and speed, 
for ten data sizes and random initial misalignments. 
The bottom right area of the plot, dominated by FGA, reflects the most efficient marks. 
} 
\label{fig:RuntimeAccuracy} 
\end{figure}
\hl{To evaluate the runtime versus accuracy of FGA,} we take a clean \textit{bunny} 
with ${\approx}35k$ points and subsample it with ten increasing subsampling factors. 
\hl{Ten $(\mathbf{X, Y})$ pairs} are obtained applying random rigid transformations on each $\mathbf{Y}$. 
Fig.~\ref{fig:RuntimeAccuracy} illustrates 
the computational throughput on a CPU in frames per second and the accuracy as RMSE of FGA 
(for $\theta=0.5, 0.6, 0.7$) against other methods, \textit{i.e.,} 
CPD~\cite{MyronenkoSong2010}, GMMReg~\cite{GMMReg1544863}, 
FilterReg~\cite{Gao2019}, FGR~\cite{FGRECCV16}, RANSAC~\cite{Rusu2009}, 
point-to-point ICP~\cite{BeslMcKay1992} and GA~\cite{Golyanik2016GravitationalAF}. 
FGA ranks top in terms of its \hl{computational throughput and accuracy} for large point \hl{set sizes}. 
Only FilterReg~\cite{Gao2019} and FGR~\cite{FGRECCV16} rival our method on a small 
subset of cases, see Fig.~\ref{fig:RuntimeAccuracy}-(B) for a bivariate correlation
plot for RMSE and the registration speed. 
Thanks to the full data parallelization of FGA, its GPU version runs ${\sim}100$ times 
faster than its CPU version \hl{and also outperforms in speed all other tested CPU versions.} 
\hl{Note the GPU version of FGA corresponds entirely to the CPU version, and the 
negligible discrepancy in the RMSE is due to differences in floating-point calculations 
between CPU and GPU and the possible \hltt{different} number of iterations.}

\begin{figure}[!ht]
 \begin{center}
   \includegraphics[width=0.75\linewidth]{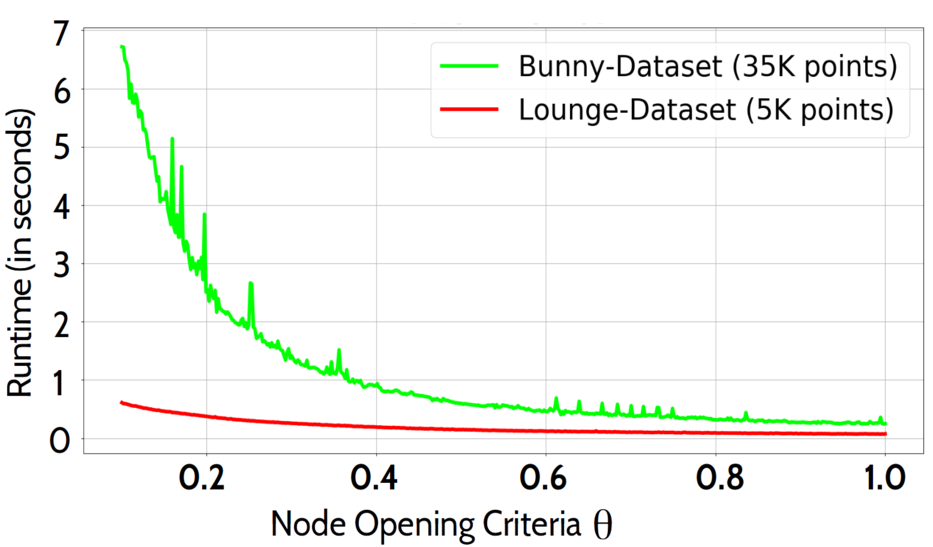}  \end{center}
  \caption{Runtime (in seconds) of FGA for increasing values of the node opening criteria 
  $\theta$ in FGA.} \label{fig:ThetaToRuntime}
\end{figure}

The multi-pole acceptance criteria $\theta$ reduces \hl{the number of} BH tree traversals
with its increasing value from $0$ to $1$. 
The effect of increasing $\theta$ to reduce the computational time of FGA is reflected 
in Fig.~\ref{fig:ThetaToRuntime}. 
The reason \hl{for the} speed-up is that the \hl{information about the nodes 
(positions and masses)} at higher depths are being summarized by 
the nodes of the same type at a lower depth. 
In the range $\theta \in (0, 0.6]$, the loss of accuracy in \hl{the} gravitational force 
approximation is negligible. 
\hl{Note that} for some datasets with a high percentage of noise and data discontinuities, 
choosing a higher range of $\theta \in [0.7, 1]$ is not suitable to trade \textit{speed} 
for \textit{accuracy gain}. 
\hl{Note that a spiking effect is observed on the runtime curve for ${\approx}35k$ 
points in Fig.~\ref{fig:ThetaToRuntime}.} 
\hl{The reason is that for larger point sets,} the number of opened nodes does not 
\hl{continuously} change (increase or decrease) 
for a continuous change on the $\theta$ value. \hl{Moreover, due to the same reason,} the alignment process can require \hl{a} different
number of iterations to converge. 

\vspace{-0.25cm}
\subsection{Partially Overlapping Depth Data}\label{subsec:Partially_OverlappingData} 
\begin{figure}[!h]
\begin{center}
   \includegraphics[width=0.99\linewidth]{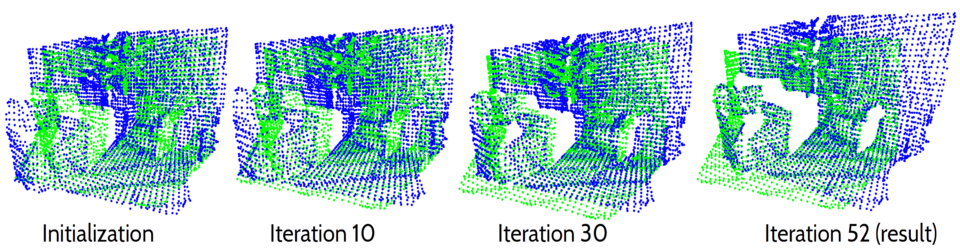}
\end{center}
   \caption{A sample result of FGA tested with two frames (separated by four 
   consecutive frames) with moderate partial overlaps from the Stanford \textit{lounge} dataset \cite{Choi_2015_CVPR}.} 
\label{fig:LoungeDataset_FGA}
\end{figure}
\begin{table}[!h]
\setlength{\tabcolsep}{3pt}
 \begin{center}
\begin{tabular}{|l|c|c|c|c|}
      \hline
      & \shortstack{Angular Err$(\varphi)$
      \\ $\varphi_{avg.}, \varphi_{min}, \varphi_{max}$}
      & \shortstack{success$\,\%$
      \\ ($\varphi < 4^\circ$)}
      & \shortstack{success$\,\%$
      \\ ($\varphi < 3^\circ$)}
      & \shortstack{success$\,\%$
      \\ ($\varphi < 2^\circ$)}  \\ \cline{1-5}
      \textbf{FGA (ours)}	      			& 2.74, {\textcolor{orange}{\bf 0.071}}, $\,19.61$ & {\textcolor{orange}{\bf 85.8\%}} &{\textcolor{orange}{\bf 75.8\%}} &{\textcolor{orange}{\bf 65\%}}\\
      \hline
      \textbf{GA}~\cite{Golyanik2016GravitationalAF} 	& $2.85,\,0.148,\,15.73$ 	  & $77.2\%$ 	      &$66\%$		 &$58.7\%$\\
      \hline
      \textbf{CPD}~\cite{MyronenkoSong2010}	      	& $3.21,\,0.075,\,32.01$ 	  & $72\%$   	      &$60.8\%$	         &$51.7\%$\\
      \hline
      \textbf{ICP}~\cite{BeslMcKay1992}      		& $3.32,\,0.284,\,14.92$ 	  & $72.8\%$ 	      &$57.5\%$	         &$46.5\%$\\
      \hline
      \textbf{GMMReg}~\cite{GMMReg1544863}	      	& {\textcolor{orange} {\bf 1.15}}, $0.072,\,16.61$    &{\textcolor{orange}{\bf 91.2\%}} 	&{\textcolor{orange}{\bf 90.1\%}}   &{\textcolor{orange}{\bf 90\%}}\\
      \hline
      \textbf{FilterReg}~\cite{Gao2019}    		& 2.67, 0.260, {\textcolor{orange} {\bf 14.91}}       & $85\%$ 	      &$74\%$		 &$59\%$\\
      \hline
      \textbf{FGR}~\cite{FGRECCV16}    		& 3.267, 0.214, 16.09 	  & $78.8\%$ 	&$63.5\%$	&$26.7\%$\\
      \hline
      \textbf{RANSAC}~\cite{Rusu2009}    		& 3.498, 0.189, 31.80 	  & $71.33\%$ 	&$68.77\%$	&$39.1\%$\\
      \hline
\end{tabular}
\end{center}
\caption{The success rate of compared methods for three upper bounds 
($4^\circ, 3^\circ, 2^\circ$) on angular deviation from ground truth after registration.} \label{table:LoungeDataset_PartialScan}
\end{table}
In \hl{RGB-D based SLAM}, globally-optimal rigid alignment \hl{provides} camera trajectories 
by mapping partial scenes. 
\hl{For our} quantitative evaluation, we choose the first $400$ frames of depth data 
from Stanford \textit{lounge}~\cite{Choi_2015_CVPR} dataset. 
We next perform registration on every fifth frame (see Fig.~\ref{fig:LoungeDataset_FGA}) 
after down-sampling those to ${\approx}5000$ points each and report the final Euler angular 
deviation $\varphi$ (\hl{no prior matches are used in this experiment}).   
Three different success rates of FGA are measured as the percentages of total experimental 
outcomes when $\varphi$ is below three different cut-off levels --- $4^\circ, 3^\circ,\,\, 
\text{and} \,\, 2^\circ$, respectively. 
\hl{Moreover, we report the average, minimum and maximum angular errors denoted by 
$\varphi_{avg.}, \varphi_{min}$ and $\varphi_{max}$, respectively, see 
Table~\ref{table:LoungeDataset_PartialScan}.} 
On CPU, FGA takes around $1.2$ seconds on average until convergence (in $30-60$ iterations) 
and performs as \hl{the second most accurate method} after GMMReg~\cite{GMMReg1544863}. 
\hl{Note that FGA achieves the lowest $\varphi_{min}$ among all compared methods}.
%

\vspace{-0.2cm}
\subsection{Pairwise Registration of Indoor Scenes}\label{subsec:3DMatch}
\begin{figure*}[!h]
\begin{center}
   \includegraphics[width=0.99\linewidth]{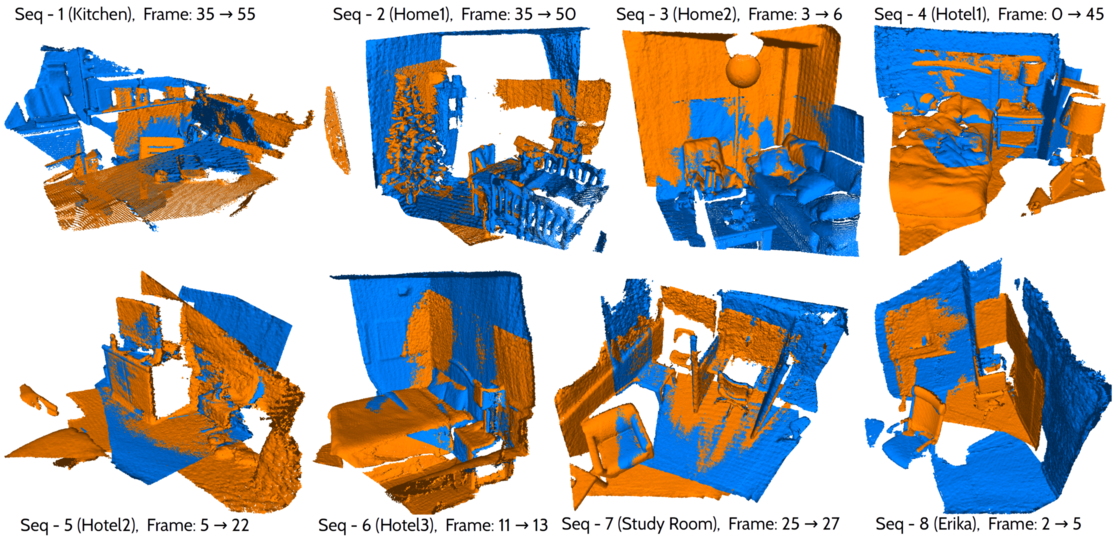}
\end{center}
   \caption{Sample pairwise registration results of our method on all eight different 
   test scenes in 3DMatch~\cite{zeng20163dmatch} dataset. 
            The input point set pairs exhibit minimal overlaps and irregular cropped areas. 
      The template $\mathbf{Y}$ is colored blue and the reference is colored orange.
      Results of FGA named by \hl{the source-to-target frames} ($i\,\rightarrow\,j$) 
   are best viewed in color. 
      } 
\label{fig:3DMatchDataset_FGA}
\end{figure*}
\begin{figure}[!h]
\begin{center}
   \includegraphics[width=0.99\linewidth]{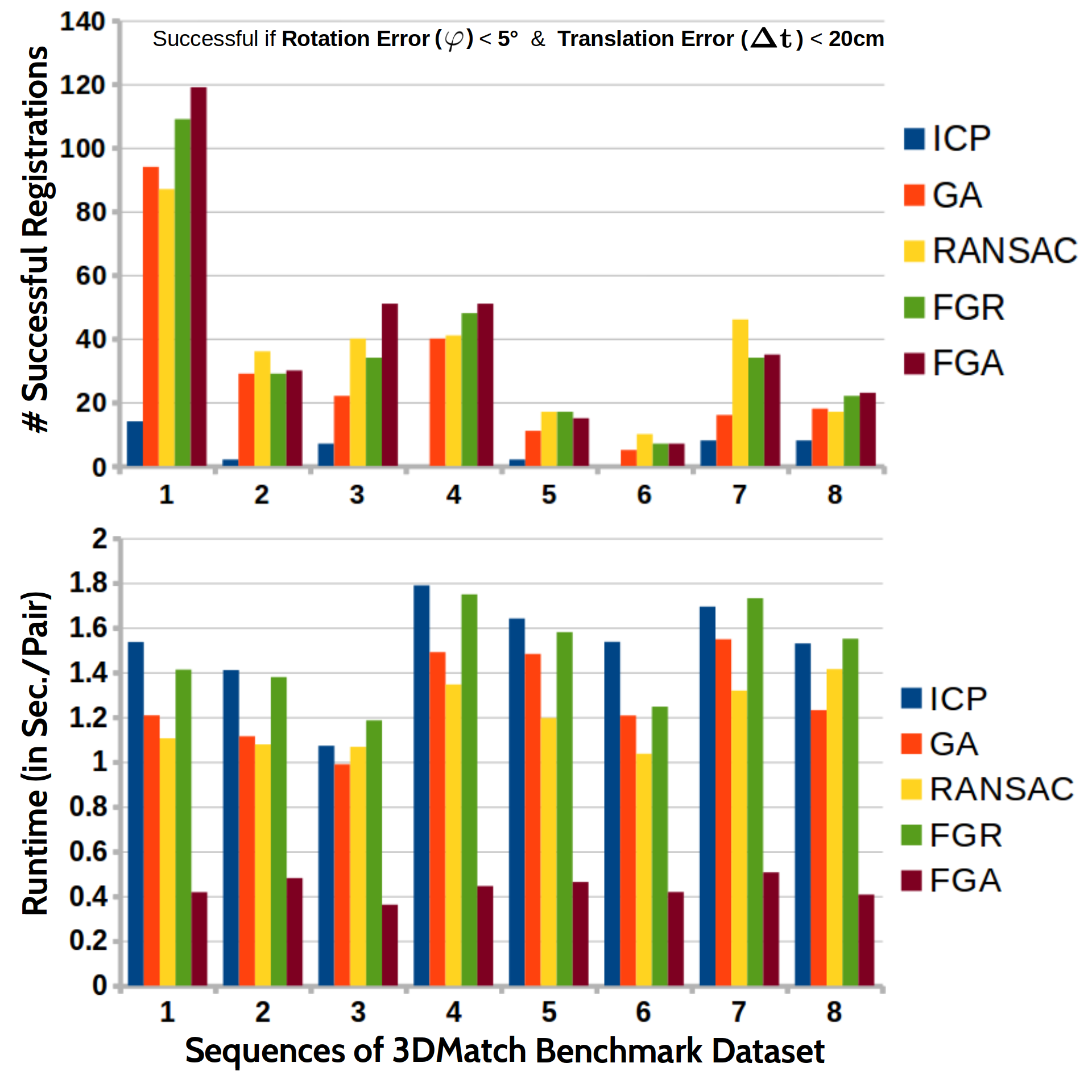}
\end{center}
   \caption{
   Our FGA is tested on eight evaluation sets of 3DMatch~\cite{zeng20163dmatch} dataset. 
      We define a strict measurement (\textit{i.e.,} when the angular error is less than $5^\circ$ 
   and translational error is less than $20$ centimeters) to count successful registrations. 
      \textit{Top plot:} The success rate of FGA is compared against the previously well-performing methods 
   --- FGR~\cite{FGRECCV16}, RANSAC~\cite{Rusu2009} and two other baseline methods ---  
   GA~\cite{Golyanik2016GravitationalAF} and ICP~\cite{BeslMcKay1992}. 
      \textit{Bottom plot:} FGA records the fastest runtime of ${\sim}0.423$ seconds, including 
   mass computation via SPM function, on average per a scan pair.} 
\label{fig:3DMatchDataset_FGA_Eval} 
\end{figure}

\begin{figure}[!h]
\begin{center}
   \includegraphics[width=0.95\linewidth]{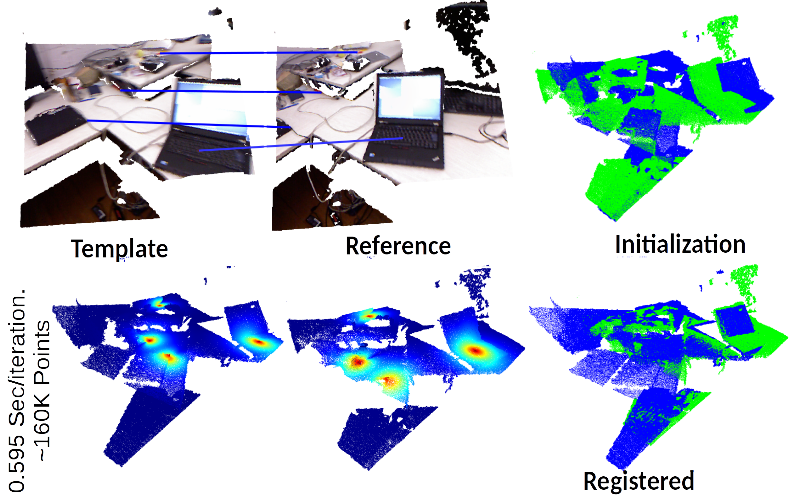} 
\end{center}
   \caption{FGA applied on \textcolor{blue}{\textit{template}} and \textcolor{green}{\textit{reference}}
   point clouds (frames $1$ and $10$ from \textit{Freiburg}~\cite{sturm12iros} dataset). 
      \textit{Top row}: input images with the prior correspondenes and the initialization on the right. 
   \textit{Bottom row}: Our SPM ($\mathbf{S}$) function radially distributes weights around \textit{four} different prior landmarks 
   (highlighted by blue lines), and the alignment on the right. 
         } 
\label{fig:TUM_ssRegistration}
\end{figure}
\begin{figure*}[!h]
 \begin{center}
   \includegraphics[width=0.99\linewidth]{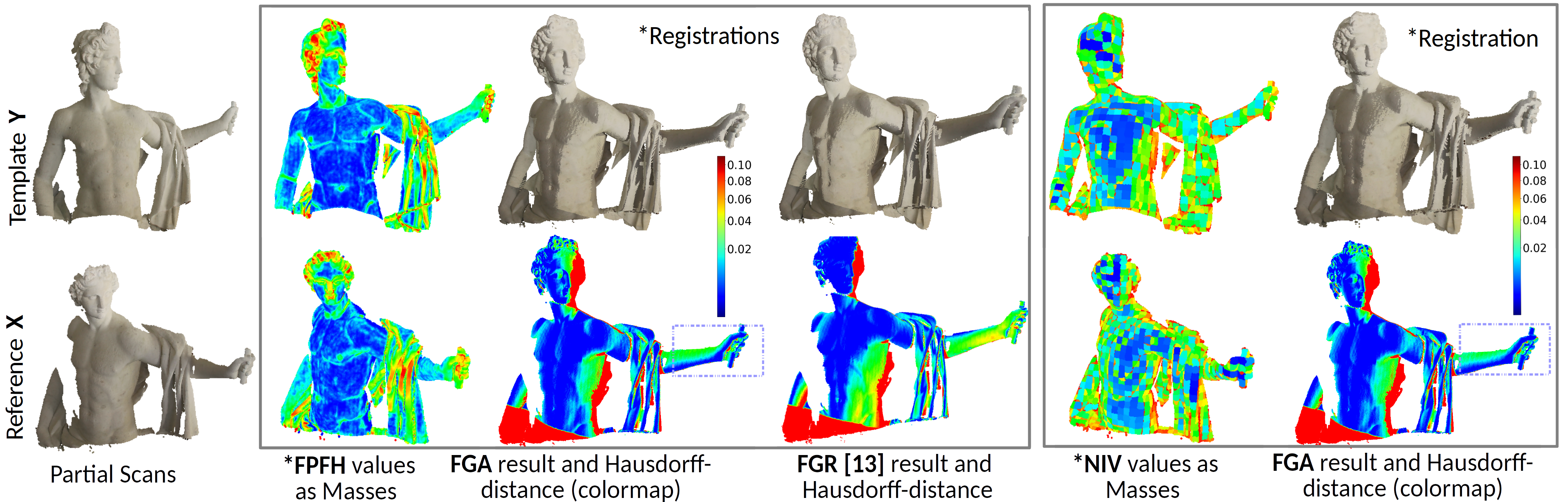}
  \end{center}
  \caption{Partial scans of the Apollo statue aligned using FGA, with the point masses distributed using feature values. 
  We use our smooth-particle mass mapping as a per-point feature and FPFH~\cite{Rusu2009} as an alternative option to demonstrate the performance of FGA compared to 
  FGR~\cite{FGRECCV16} (the second-best performing method). The blue$<$green$<$red color scheme shows the Hausdorff distance on the template's surface after the alignment.} 
 \label{fig:FeatureBased_Alignment} 
\end{figure*} 

The 3DMatch~\cite{zeng20163dmatch} benchmark dataset contains eight sequences  
of minimally-overlapping partial scan pairs of indoor scenes. 
It is thus highly challenging for unsupervised point set alignment methods. 
CPD~\cite{MyronenkoSong2010}, ICP~\cite{BeslMcKay1992}, GMMReg~\cite{GMMReg1544863} and
FilterReg~\cite{Gao2019} all perform poorly on partial data and particularly on 3DMatch. 
\hl{Furthermore, subsampling of 3DMatch scans with a voxel size of $3$ cm results in 
point clouds with ${\sim}15k$ points which is still computationally expensive for these methods.}  
The evaluation includes $506$, $156$, $207$, $226$, $104$, $54$, $292$ and $77$ challenging 
pairs from the 3DMatch sequences. 
Fig.~\ref{fig:3DMatchDataset_FGA} depicts sample registration results of FGA on pairs of 
scans from all eight sequences. 
Fig.~\ref{fig:3DMatchDataset_FGA_Eval} shows the total number of successful registrations using 
FGA and its runtime compared to other benchmark methods on this dataset. 
We have also chosen a strict upper bound on rotational error (<$5^\circ$) and translational 
error (<$20cm$) to define the success parameter. 
Our FGA outperforms all compared methods in the overall number of successful registrations 
over all eight sequences. 
In the runtime, FGA outperforms other methods by the factor ranging from ${\approx}2.5$ to ${\approx}4$. 
FGR~\cite{FGRECCV16} and RANSAC~\cite{doi:10.1111/cgf.12446, Rusu2009} are ranked second and 
third in the success rate and outperform other compared methods except for FGA.

\noindent\textbf{Using Prior Correspondences} $\mathbf{(1)}.$ A qualitative evaluation is 
also performed by testing FGA on RGB-D 
\textit{Freiburg}~\cite{sturm12iros} dataset picking \hl{frames $1$ and $10$} as 
$\mathbf{Y}$ (${\approx}220k$ points) and $\mathbf{X}$ (${\approx}240k$ points). 
$Four$ landmark points --- 
(i) \textit{\lq red button\rq}$\,$of the laptop, (ii) tip of a \textit{\lq red pen\rq},
(iii) corner of a \textit{\lq document folder\rq}$\,$and (iv) a  point near \textit{\lq red apple\rq}$\,$
--- are manually selected. They can perhaps be non-exact in their positional accuracy. 
The scenes are \hl{moderately}   
overlapping. The RBF values are higher around the landmarks. The results and the runtime of FGA 
are shown in Fig.~\ref{fig:TUM_ssRegistration}.

\noindent\textbf{Using Feature-Based Particle Masses} $\mathbf{(2)}.$ \hl{The next experiment} 
emphasizes that FGA can use different types of point-based and geometry-based features as input 
weights (point masses). 
\hl{We set} products of the masses of \hl{the} template and reference points \hl{as} directly 
proportional to their gravitational forces of attraction. 
We choose (a) values of FPFH~\cite{Rusu2009} and (b) the values of our SPM function 
\hl{(this is equivalent to the NIV values with no landmarks)} as the masses of structured-light 
scan point clouds with $159k$ and $145k$ points, respectively. 
In both cases, FGA registers all input scans accurately.  There is a marginally higher accuracy \hl{when} using NIV values as point features. 
We also compare FGA and FGR on Apollo scans with partial overlaps, see Fig.~\ref{fig:FeatureBased_Alignment}. 
The Hausdorff distances sampled on $\mathbf{X}$ show a higher proportion of blue color on the template's surface.

\vspace{-0.2cm}
\subsection{Robustness against Data Disturbances}\label{subsec:Robustness_Against_Data_Disubances} 
\begin{figure*}[!h]
 \begin{center}
   \includegraphics[width=0.99\linewidth]{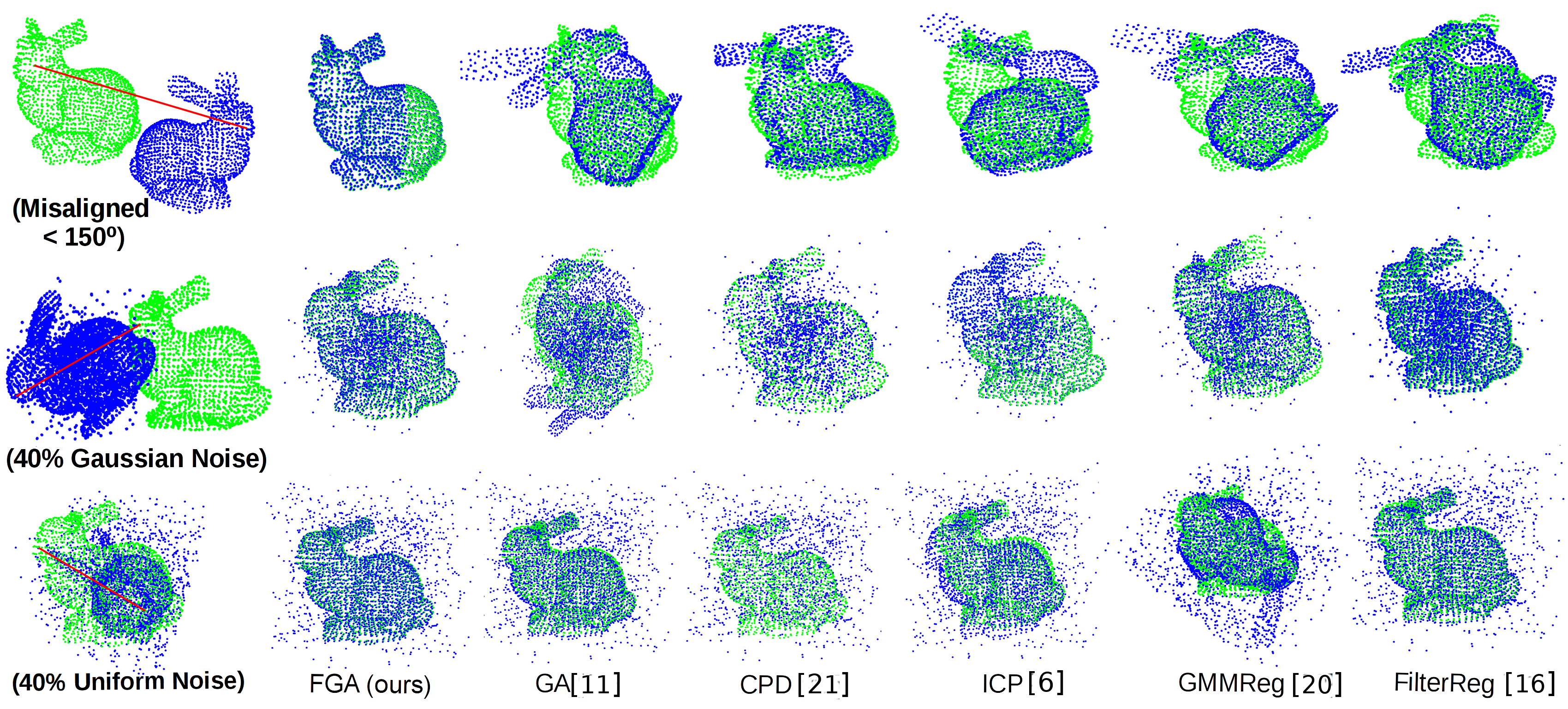}
  \end{center}
  \caption{ 
Exemplary registration results in the experiment with the data disturbances 
(applying large misalignment, $40\%$ of Gaussian noise, and $40\%$ of uniformly distributed noise 
on the template $\mathbf{Y}$). 
Our FGA uses one prior match in this visualization and outperforms competing methods
also without prior correspondences. 
See Table~\ref{table:BunnyRobustnessEvaluation} for complete quantitative results 
(also without prior correspondences). 
} 
\label{fig:BunnyRobustnessEvaluationQualitative} 
\end{figure*}

To evaluate the robustness of FGA against \hl{different disturbing effects}, 
we take a clean \textit{bunny} \hl{with $1889$ points} and add \hl{to it} \textbf{(a)} 
$40\%$ (of $M$) random Gaussian, \textbf{(b)} $40\%$ (of $M$) random uniformly distributed noise. 
\hl{We next} \textbf{(c)} transform \hl{the samples} randomly with angular deviations 
$\varphi_x, \varphi_y,$ and $\varphi_z$ where all are $\in\mathcal{U}(0, \frac{3\pi}{4})$. 
We prepare $100$ such independent test samples for each of the three cases and report 
success rate when RMSE value is ${<}0.01$, 
with average speed (fps) comparison in Table~\ref{table:BunnyRobustnessEvaluation} and visualizations in Fig.~\ref{fig:BunnyRobustnessEvaluationQualitative}. 
FGA performs as the second-best method after CPD~\cite{MyronenkoSong2010} in the presence of noise. 
In extreme noisy input scenarios (when the amount of noise is ${>}60\%$ of input data), 
the NIV ($\mathbf{N}$) measure 
can be constant (\hbox{\eg $\mathbf{N}(\mathbf{X})$ and $\mathbf{N}(\mathbf{Y}) = 1$}). 
FGA is more efficient and \hl{accurate} in registering \hl{substantially} misaligned data when 
\hl{only} a few landmark correspondences are available. 
In this case, even without any landmark correspondences, FGA has the highest success rate of $62\%$, 
whereas FilterReg's performance is the worst with $28\%$.
\begin{table}[!t]
\setlength{\tabcolsep}{2pt}
   \begin{center}
\begin{tabular}{|l|c|c|c|}
      \hline
	& \shortstack{misalign$(< 150^\circ)$
	\\ (success$\%$, fps)}
	& \shortstack{$40\%\,\mathcal{U}$ noise
	\\ (success$\%$, fps)}
	& \shortstack{$40\%\,\mathcal{G}$ noise
	\\ (success$\%$, fps)} \\ \cline{1-4}
	\textbf{FGA (ours)}				& {\textcolor{orange}{\bf 62\%, 130.1}} & $69\%,\,121.3$ & $78\%,89.1$		\\
	\hline
	\textbf{FGA (ours)/}$1^\dagger$				& {\textcolor{orange}{\bf 66\%, 126.1}} & $63\%,\,73$    & $72\%,78$		\\
	\hline
	\textbf{FGA (ours)/}$3^\dagger$				& {\textcolor{orange}{\bf 82\%, 125}}   & $79\%,\,73$	 & {\textcolor{orange}{\bf 81\%}}, 78	\\
	\hline
	\textbf{GA}~\cite{Golyanik2016GravitationalAF}& $44\%,\,8$	 	  & $66\%,\,6.5$		& $75\%,\,8.1$		 \\
	\hline
	\textbf{CPD}~\cite{MyronenkoSong2010}	  	& $59\%,\,4.9$ 	 	  & {\textcolor{orange}{\bf 95\%}}, 3.5	& {\textcolor{orange}{\bf 93\%}},\,4.98 \\
	\hline
	\textbf{ICP}~\cite{BeslMcKay1992}  	& $40\%,\,113$ 	 	  & $6\%,\,39.1$		& $62,\%,25$		 \\
	\hline
	\textbf{GMMReg}~\cite{GMMReg1544863}		& $32\%,\,18.1$  	  & $55\%,\,29$      		& $49\%,\,39$		 \\
	\hline
	\textbf{FilterReg}~\cite{Gao2019} 		& {\bf 28\%},\,72.2  & $62\%,\,46.8$& $51\%,\,58.2$		 
	\\ \hline
\end{tabular}
\end{center}
\caption{Evaluation on \textit{bunny} under Gaussian and uniform noise and large misalignment. 
``$^{\dagger}$'' indicates the number of prior matches for FGA. 
} 
\label{table:BunnyRobustnessEvaluation}
\end{table}

\vspace{-0.25cm}
\subsection{LIDAR Odometry}\label{subsec:LiDAR_Odom_Exoeriments}
\begin{figure*}[!h]
 \begin{center}
   \includegraphics[width=0.99\linewidth]{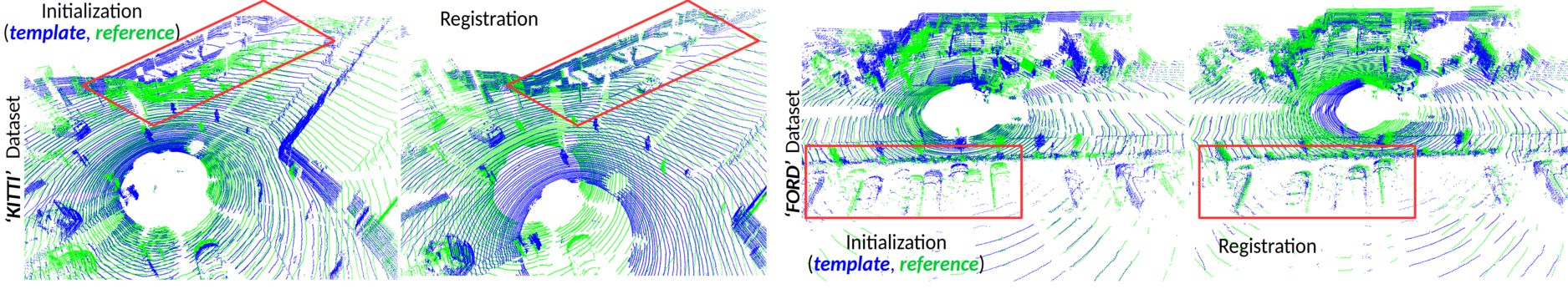}
  \end{center}
  \caption{Registration results of FGA on KITTI~\cite{Geiger2013IJRR} and Ford
  Campus Vision~\cite{Pandey:2011:FCV:2049736.2049742} datasets without subsampling. \textit{Left:}
  \hl{frames $1$ and $20$} as $\mathbf{Y}$ and $\mathbf{X}$,
  respectively ($M + N {\approx} 245k$ points) from $2011\_09\_26\_drive\_0005\_sync$ driving sequence. 
    \textit{Right:} \hl{frames $1000$ and $1010$} as $\mathbf{Y}$ and $\mathbf{X}$ 
  ($M + N {\approx} 150K$ points). 
    In both cases, no landmarks are available. 
    }
\label{fig:LIDAR_Registration}
\end{figure*}
\begin{figure}[!h]
 \begin{center}
   \includegraphics[width=0.99\linewidth]{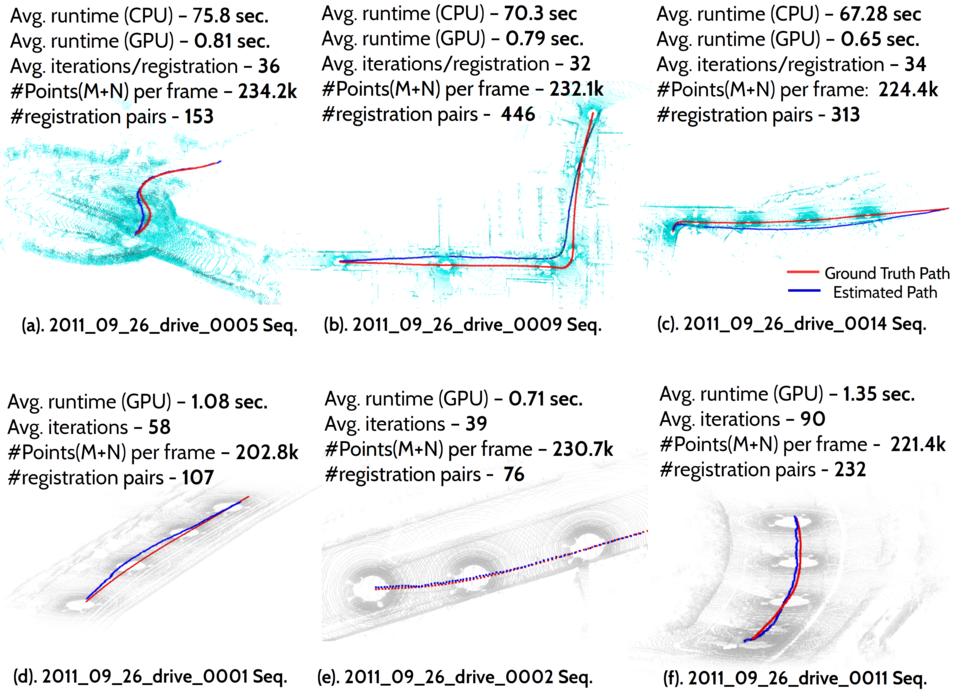}
  \end{center}
  \caption{Consecutive frame-wise registrations of LiDAR point clouds from \textit{six} 
  different driving sequences 
  of the KITTI-RAW~\cite{Geiger2013IJRR} dataset \hl{by our FGA}. 
    The ground-truth paths are compared with the estimated paths, which reflects minimum drift and 
  transformation error. 
    In the \textit{first} row, the driving sequences have more turns in the trajectories 
  of sensor-carrying vehicle. In contrast, the vehicle trajectories in the \textit{second}  
  row have relatively more \hl{pronounced} rectilinear motion than in the \textit{first} row. 
  } 
\label{fig:LIDARPath}
\end{figure} 
\hl{LiDAR data exhibits varying point sampling densities at the near and far fields.} 
\hl{Already} in the early research, \textit{Occupancy Grid}~\cite{Elfes:1989:UOG:68491.68495} 
mentioned cases where robotic navigation has to deal with differently tessellated spatial information. 
Our NIV measure is especially helpful in such scenarios as it plays the role of a 
density map to spread out weights across the point clouds. 
We choose the LiDAR dataset available from the \textit{KITTI}~\cite{Geiger2013IJRR} and the 
\textit{Ford Campus Vision}~\cite{Pandey:2011:FCV:2049736.2049742} benchmark. 
No prior point correspondences are used in these experiments. 
Hence, our SPM function is equal to the NIV values on the point clouds. 

Fig.~\ref{fig:LIDAR_Registration} demonstrates results of FGA for two navigation scenarios where a car 
is either --- (i) taking some turn (in KITTI dataset) as orientation change is dominant, 
or (ii) moving forward (in \textit{Ford} dataset) as translation part is dominant. 
FGA correctly aligns the predominant scene components such as cars (marked by 
\textit{red boxes}) and buildings. 
Note how FGA ignores \hl{the points concentrated} at the central part
and rotates $\mathbf{Y}$ towards $\mathbf{X}$ in the KITTI test or translates 
\hl{the template point cloud} in the forward direction to align it with 
$\mathbf{X}$ on the \textit{Ford} \hl{dataset}. 
Our method without the NIV map registers incorrectly. 
Other methods, except FilterReg, FGR, and RANSAC, cannot process such large point clouds. 
Subsampling here can result in removing a few but necessary data (\eg the number of points 
representing a pedestrian is only ${\sim}100$).

\begin{figure}[!t] 
 \begin{center} 
   \includegraphics[width=0.99\linewidth]{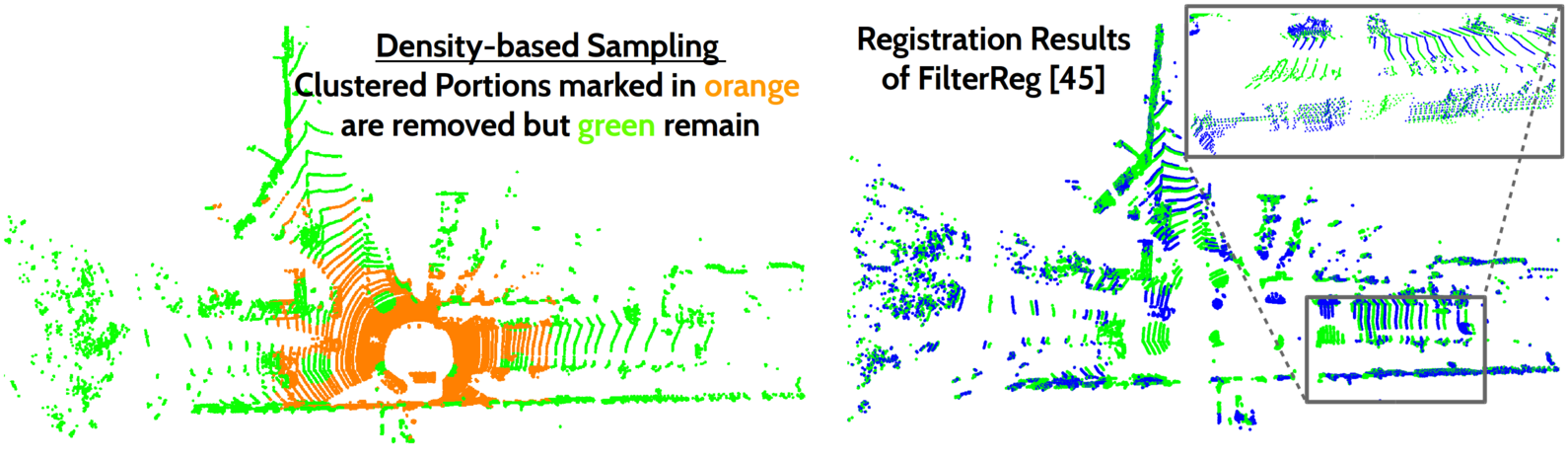} 
  \end{center} 
  \caption{\textit{On the right}: Density-based subsampling of input LiDAR point clouds 
  which keeps the salient points (\textit{e.g.,} the points representing the cars, facades and pedestrians)  
  and removes mainly the points in the central area with a large amount of clutters. 
  \textit{On the left}: This careful sampling allows other robust 
  registration methods, \textit{e.g.,} FilterReg~\cite{Gao2019} to find globally-optimal 
  correspondences.} 
\label{fig:KITTI_SAMPLING} 
\end{figure}

The CPU and GPU versions of FGA \hl{run} on \textit{six} different 
\textit{KITTI}~\cite{Geiger2013IJRR} sequences resulting in a total of $1327$ 
experiments to obtain the sensor trajectories. 
We apply consecutive frame-to-frame alignments, where each alignment \hl{involves} 
approximately $220k$ points. 
Fig.~\ref{fig:LIDARPath} illustrates the ground-truth path in red and our estimated 
path in blue for three of the sequences with minimum sensor drifts. 
\hl{FGA requires $70$ seconds on CPU and $0.7$ seconds on GPU on average to align a pair 
of frames (the speedup of two orders of magnitude).} 
If only the sensor's forward view is considered or points on the ground are removed, 
the input data size \hl{is} reduced by ${\approx}40\%$. 
FGA can process $8{-}10$ frames per second for such input data size. 
Compared to FGA, other methods 
--- 
GA~\cite{Golyanik2016GravitationalAF}, 
CPD~\cite{MyronenkoSong2010}, point-to-point ICP~\cite{BeslMcKay1992} and GMMReg~\cite{GMMReg1544863} 
cannot process \hl{input point clouds} of large size, except FilterReg~\cite{Gao2019}, 
FGR~\cite{FGRECCV16} and RANSAC~\cite{Rusu2009}.

\begin{figure}[!t] 
 \begin{center}
   \includegraphics[width=0.99\linewidth]{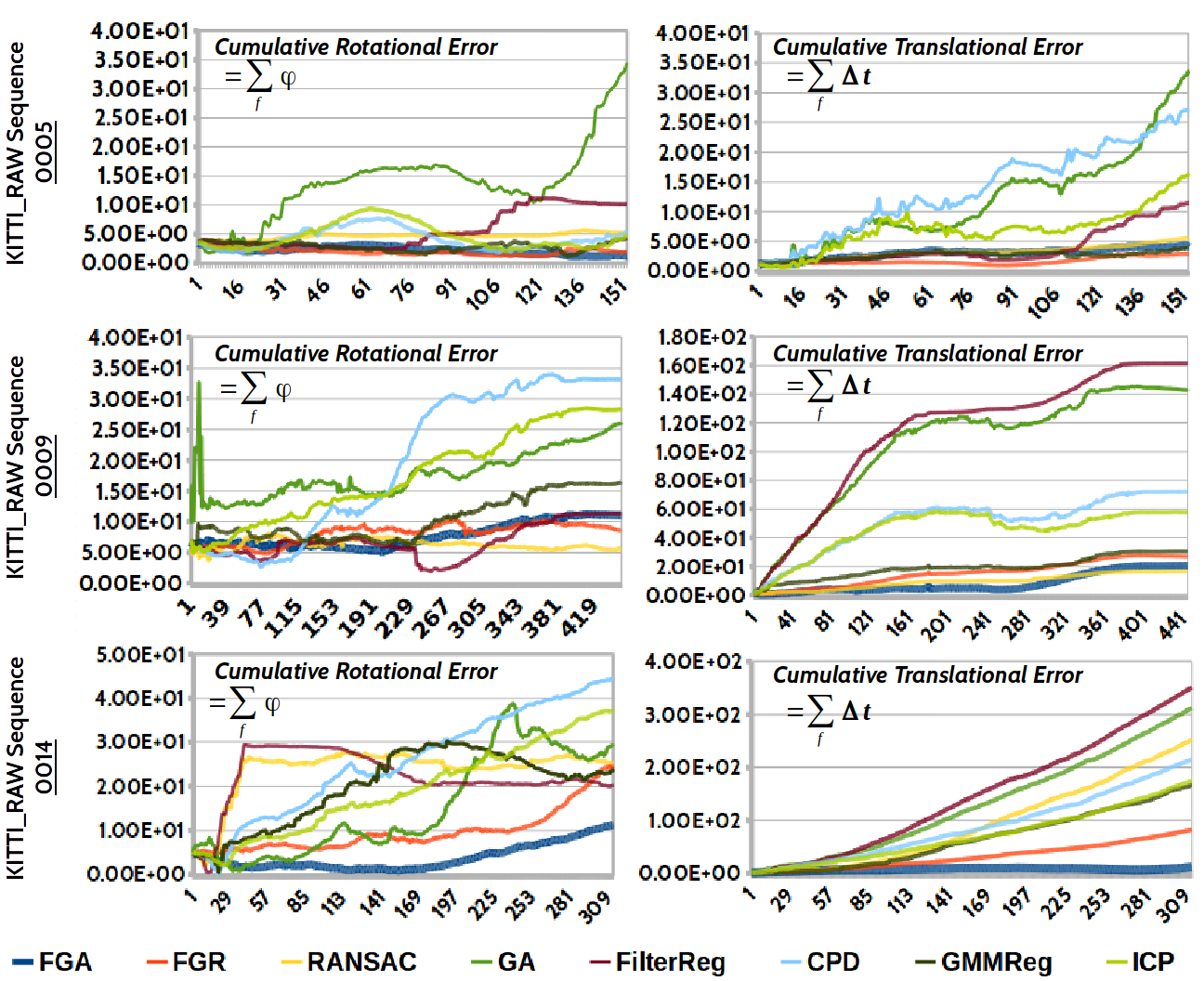} 
  \end{center}
  \caption{Quantitative evaluation of several methods on the KITTI-RAW dataset. 
        Plots show cumulative \hl{rotational and translational errors} accumulated over 
  pairwise registrations of all consecutive frames. 
        Cumulative translational errors remain lowest across the frames for a given sequence 
  \hl{in the case of our FGA}. 
    Only FGR~\cite{FGRECCV16} and RANSAC~\cite{Rusu2009} show either comparable or higher  
  accuracy over a few short subsequences. 
    } 
\label{fig:KITTI_PATH_QUANTITATIVE}
\end{figure}

Thus, to evaluate the competing methods (CPD, ICP, RMMReg and RANSAC), 
we need inputs of substantially small sizes. 
This can be achieved by either extracting feature descriptors from large inputs or subsampling 
those with large subsampling factors.  
Hence, we use density-based sampling of the input point clouds as shown in 
Fig.~\ref{fig:KITTI_SAMPLING}, \textit{i.e.,} we remove parts of the point 
clouds in which point density exceeds a threshold. 
For the sake of speed, this sampling is done naively using regular volumetric non-overlapping bins. 
In this evaluation, the transformation errors resulting from the pairwise 
registrations of the LiDAR frames using any method will be lower than that 
would result by using a simple uniform sampling. 
The plots in Fig.~\ref{fig:KITTI_PATH_QUANTITATIVE} quantify the \textit{cumulative sum} of 
the angular errors $(\varphi)$ and translation errors $(\mathbf{\Delta t})$ between estimated 
and ground-truth poses of the sensor for three different driving sequences from the KITTI-RAW data. 
It shows that the compared methods can estimate the sensor orientations (in degree) 
reasonably close to the ground truth but the translation errors (in centimeters) 
reflect that FGA, FGR and FilterReg are the standouts among them. 
The cumulative translation errors across all the three driving sequences are minimum when using FGA. 
We run the frame-to-frame registration using FGA on a GPU for all sequences of KITTI RAW dataset. 
Estimated vehicle trajectories for all other sequences are compared against the ground-truth path in Fig.~\ref{fig:LIDARPath}. 
The estimated paths are close 
to the ground-truth paths without applying any refinement or loop closure methods. 
\begin{figure*}[!ht] 
 \begin{center} 
   \includegraphics[width=0.99\linewidth]{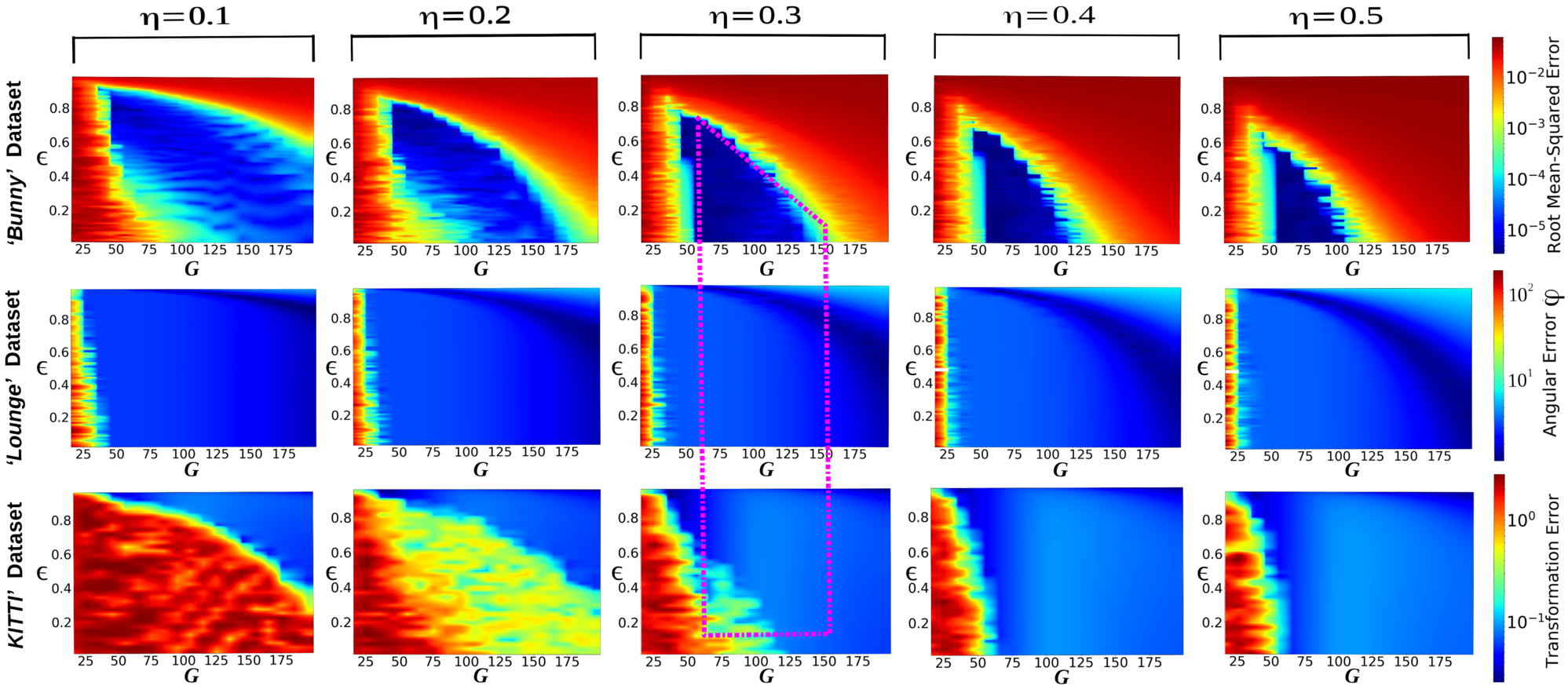} 
  \end{center} 
  \caption{Selecting best FGA parameters with a grid search method on $G,\epsilon$ 
  and $\eta$ (slices for $\eta = 0.1, 0.2, 0.3, 0.4$ and $0.5$). 
    These parameters can be set from the common and wide area (value ranges) marked by the polygon in the middle. 
      The parameter settings for our experiments are drawn from this region. 
    Another figure like this would also draw the same conclusion on datasets scaled within other ranges (\eg $[-1, 1]$). 
    } 
\label{fig:ParamSettingsAnalysis}
\end{figure*}

\vspace{-0.35cm}
\subsection{Parameter Selection}\label{subsec:OptimalParamChoices} 
Three main FGA parameters, \textit{i.e.,} $G,\epsilon,$ and $\eta$, account for the 
convergence basin of the objective function \eqref{eqn:GPE_Energy_GA}. 
To determine their best values, we prepare a regular 3D grid covering $G \in [1, 200],\, \epsilon \in [0.1, 0.5],$ 
and $\eta\in[0.1, 0.5]$, with the step sizes of $1, 0.1$ and \hl{$0.1$, respectively}. 
FGA is then executed for $10^4$ grid cells for each of the three datasets --- \textit{bunny, lounge} and \textit{KITTI}. 
In Fig.~\ref{fig:ParamSettingsAnalysis}, the color indicates RMSE between 
registered pairs of the \textit{bunny}. 
For \textit{lounge} and \textit{KITTI} data, the color denotes the final Euler angular error 
in Eq.~\eqref{eqn:GroundTruthAngularDev} and the total transformation error in Eq.~\eqref{eqn:Total_TransError}, 
respectively. The plots in Fig.~\ref{fig:ParamSettingsAnalysis} show that the parameters can be set 
in a wide interval for all tested datasets. 
The polygon area in the plot marks a wide and common range of parameters which can be used 
to obtain minimal alignment errors.

\vspace{-0.25cm} 
\section{Discussion}\label{sec:discussion} 
All in all, our carefully-designed and extensive experiments confirm the effectiveness of the gravitational 
point set registration paradigm, also in the context of the recent advances in point set alignment as of 2020. 
Especially with automatically-extracted features which are mapped to the point masses, 
the proposed method either performs on par with the respectively best method in a given 
scenario  or surpasses the alignment accuracy of all compared methods, as shown in 
Secs.~\ref{subsec:Partially_OverlappingData}--\ref{subsec:3DMatch}. 
From the experiments in Secs.~\ref{subsec:Speed_AccuracyAnalysis} and 
\ref{subsec:LiDAR_Odom_Exoeriments}, we also see that FGA is well suitable for 
parallelization on a single GPU. 
It thus can be used in applications requiring interactive system response (\textit{e.g.,} 
autonomous driving). 
From the correlation plot in Fig.~\ref{fig:RuntimeAccuracy}-(B), we see that FGA does not 
trade the computational speed for the accuracy as several competing methods do (\textit{e.g.,} 
CPD or GMMReg). 
Experiments in Sec.~\ref{subsec:Robustness_Against_Data_Disubances} also confirm high and 
steady robustness of FGA to noise which secures the second place for FGA in this category 
among all tested methods. 
FGA resolves larger initial misalignments compared to other tested approaches due to the 
multiply-linked character of point interactions. 
Thus, there are much fewer alignment scenarios which result in local minima compared to the 
competing techniques. 
 
The experiment for the parameter choice in Sec.~\ref{subsec:OptimalParamChoices} 
confirms that the parameters can be conveniently fixed across various datasets and scenarios. 
Thus, FGA addresses several limitations of the original GA \cite{Golyanik2016GravitationalAF} 
(\textit{i.e.,} long runtime, scenario-specific parameters) and establishes a new state of 
the art in general-purpose point set alignment. 
Along with that, FGA does not require correspondence extraction in a pre-processing step or 
local correspondence search in every iteration --- a locally-optimal alignment is achieved 
when the multiply-linked GPE is minimized. 
This GPE is known upfront and kept unchanged during the entire optimization. 

Interestingly, FGA consistently outperforms GA \cite{Golyanik2016GravitationalAF} in the accuracy and speed in all performed experiments. 
The automatically-defined boundary conditions on masses (prior matches and SPM function) 
guide the alignment procedure away from many local minima. 
We thus believe that FGA is a significant step forward in general-purpose point set 
registration techniques. 

\vspace{5pt} 
\noindent\textbf{Limitations. } Although the cues such as known landmark positions, intelligent 
use of the feature values as masses, and smooth-particle mass functions make our method 
robust, the recovered transformation is still locally-optimal. Since our method uses values 
of SPM function or other feature values as the point masses, registration accuracy depends 
upon the feature accuracy (matching precision and recall). 

\vspace{-0.1cm}
\section{Conclusions and Future Work}\label{sec:concludeGA_FGA} 
We introduce FGA --- a fast physics-based rigid point set registration method which is applicable for 
various types of data with noise and clustered outliers (such as LiDAR scans). 
While most competing methods get trapped in the local minima caused by non-uniform point sampling, our FGA 
recovers transformations which are closer to the optimal ones, thanks to the boundary 
conditions defined on masses via the SPM function. 
The acceleration policy of FGA hierarchically divides the input point clouds in a BH tree 
with local space properties, and solves the gravitational force approximation problem with 
second-order ODEs in quasilinear time. 
From the experimental results, we draw the conclusion that our method is fast (\textit{i.e.,} it can support 
interactive and real-time frame rates), accurate and robust on most of the general and more challenging datasets, especially when dealing with noise and non-uniform point sampling density.

\vspace{5pt}
\noindent\textbf{Future Work.} 
Our BH tree representation can be an efficient alternative to computationally expensive 
grid representations of point clouds for deep-learning-based alignment methods, which 
we are planning to investigate next.
Another direction is extending FGA for estimating non-rigid motion fields 
and real-time RGB-D scene flow on a single GPU, with applications to autonomous driving systems.

\bibliographystyle{IEEEtran}
\bibliography{references}
\appendix
\section{Rotation and Translation for Data Scaled in {\Large $[\mathbf{\lowercase{a,b}}]$}}
For a given input pair of \textit{template} $\mathbf{Y}$, \textit{reference} $\mathbf{X}$ and a
range $\left[a, b\right]$, the normalization of the pair to 
$(\mathbf{X,Y}) \mapsto (\mathbf{X_{\widehat{n}}},\mathbf{Y_{\widehat{n}}})$, is formalized in three steps. 
First, the mean positions of $\mathbf{X}$ and $\mathbf{Y}$, which read as  
\small
 \begin{gather}
 \mathbf{\bar{X}} = \frac{1}{N}\sum\limits_{j=1}^{N}\mathbf{X}_j; \,\,\,
 \mathbf{\bar{Y}} = \frac{1}{M}\sum\limits_{i=1}^{M}\mathbf{Y}_i;\tag{A.1}
 \end{gather}
 \normalsize
 are subtracted from the position vectors of the respective point clouds. This step shifts the center
 of the point clouds at the origin. Thereafter, the minimum and the maximum of the position vectors from the shifted 
 point clouds 
 \small
 \begin{gather}
 l = \min \left\lbrace\min\left\lbrace\cup_{i=1}^{N}(\mathbf{X}_i - \mathbf{\bar{X}})\right\rbrace, \min\left\lbrace\cup_{j=1}^{M}(\mathbf{Y}_j - \mathbf{\bar{Y}})\right\rbrace\right\rbrace,\notag\\ 
 r = \max \left\lbrace\max\left\lbrace\cup_{i=1}^{N}(\mathbf{X}_i - \mathbf{\bar{X}})\right\rbrace, \max\left\lbrace\cup_{j=1}^{M}(\mathbf{Y}_j - \mathbf{\bar{Y}})\right\rbrace\right\rbrace\tag{A.2}
 \end{gather}
 \normalsize
 are used to scale them in an arbitrary range $[a,b]$ as
 \small
 \begin{gather}\label{eqn:Normalization}
 \mathbf{X}_{\widehat{n}} = 
 \begin{bmatrix}
  \frac{(\mathbf{X}_1 - \mathbf{\bar{X}} - \mathbf{l})(b-a)}{r - l} + \mathbf{a}\\
  \vdots\\
  \frac{(\mathbf{X}_N - \mathbf{\bar{X}} - \mathbf{l})(b-a)}{r - l} + \mathbf{a}
 \end{bmatrix}^T, \,
 \mathbf{Y}_{\widehat{n}} = 
 \begin{bmatrix}
  \frac{(\mathbf{Y}_1 - \mathbf{\bar{Y}} - \mathbf{l})(b-a)}{r - l} + \mathbf{a}\\
  \vdots\\
  \frac{(\mathbf{Y}_M - \mathbf{\bar{Y}} - \mathbf{l})(b-a)}{r - l} + \mathbf{a}
 \end{bmatrix}^T,\tag{A.3}
\end{gather}
\normalsize
where $\mathbf{l} = (l,l,l)^T$ and $\mathbf{a} = (a,a,a)^T$.
FGA is applied on the normalized input data pair
$(\mathbf{X_{\widehat{n}}},\mathbf{Y_{\widehat{n}}})$. The optimal rigid transformation parameters 
$\mathbf{T^*} = \left[\mathbf{R}| \mathbf{t} \right]$, as shown in Alg.~\ref{alg:FGA_Algorithm},
register $\mathbf{Y_{\widehat{n}}}$ to $\mathbf{X_{\widehat{n}}}$ as 
$\mathbf{Y_{\widehat{n}}} \longmapsto \left[\mathbf{R}| \mathbf{t} \right] \times \mathbf{Y_{\widehat{n}}}$.
In the denormalized domain of $\mathbf{Y_{\widehat{n}}}$, the same mapping requires a denormalized translation component
\lq$\text{dnorm}(\mathbf{t})$\rq of transformation matrix $\mathbf{T^*}$ for mapping 
$\mathbf{Y}\longmapsto \left[\mathbf{R}| \text{dnorm}(\mathbf{t}) \right] \times \mathbf{Y}$. The derivation 
steps for \lq$\text{dnorm}(\mathbf{t})$\rq are:
\begin{gather}
\Rightarrow
 \left[\mathbf{R}| \text{dnorm}(\mathbf{t}) \right] \times \mathbf{Y} \equiv 
 \underbrace{\text{dnorm}(\left[\mathbf{R}| \mathbf{t} \right] \times \mathbf{Y_{\widehat{n}}})}_{
 \text{reverse of Eq.~\ref{eqn:Normalization}}}\notag\\
 \Leftrightarrow
 \left[\mathbf{R}| \text{dnorm}(\mathbf{t}) \right] \times \mathbf{Y} =
 \frac{((\left[\mathbf{R}| \mathbf{t} \right] \times \mathbf{Y_{\widehat{n}}}) - \mathbf{a})(r-l)}{(b-a)} 
 + \mathbf{\bar{Y}} + l\notag\\
 \Leftrightarrow \mathbf{R}\mathbf{Y} + \text{dnorm}(t) =
 \frac{((\left[\mathbf{R}| \mathbf{t} \right] \times \mathbf{Y_{\widehat{n}}}) - \mathbf{a})(r-l)}{(b-a)}
 + \mathbf{\bar{Y}} + l,\notag
\end{gather}
and after substituting $\mathbf{Y_{\widehat{n}}}$ by the right hand side of Eq.(A.3)
\begin{align}
\Leftrightarrow\hdots
 = \frac{\left( \mathbf{R}\left(\frac{(\mathbf{Y} - \mathbf{\bar{Y}} - \mathbf{a})(b-a)}{(r-l)} 
 + \mathbf{a}\right)  + \mathbf{t} - \mathbf{a}\right)(r-l)}{(b-a)} 
 + \mathbf{\bar{Y}} + \mathbf{l}\notag\\
 \Leftrightarrow\hdots 
 = \mathbf{RY} 
 \underbrace{-\mathbf{R\bar{Y}} - \mathbf{R}\mathbf{l} + \frac{(r-l)}{(b-a)}(\mathbf{R}\mathbf{a} + t - \mathbf{a})
 + \mathbf{\bar{Y}} + \mathbf{l}}_{\text{dnorm}(\mathbf{t})}.\notag
\end{align}
As it stands, the difference between the centroids $\mathbf{\bar{Y}}-\mathbf{\bar{X}}$ must be added to 
$\text{dnorm}(\mathbf{t})$, as the translation was computed for the normalized point sets which are centered on the origin. 
\vfill
\EOD
\end{document}